\documentclass[sigplan,10pt]{acmart}
\copyrightyear{2025}
\acmYear{2025}
\setcopyright{cc}
\setcctype{by}
\acmConference[SOSP '25]{ACM SIGOPS 31st Symposium on Operating Systems Principles}{October 13--16, 2025}{Seoul, Republic of Korea}
\acmBooktitle{ACM SIGOPS 31st Symposium on Operating Systems Principles (SOSP '25), October 13--16, 2025, Seoul, Republic of Korea}\acmDOI{10.1145/3731569.3764855}
\acmISBN{979-8-4007-1870-0/2025/10}

\usepackage{hyperref}

\usepackage{tikz}
\usepackage{amsmath}
\usepackage{listings, color}

\usepackage{filecontents}

\usepackage{tikz}
\usepackage{xcolor}
\usepackage{xspace}
\usepackage{tikz}
\usepackage{pifont}
\usepackage[english]{babel}
\usepackage{blindtext}
\usepackage{lipsum}

\usepackage[ruled,vlined,linesnumbered]{algorithm2e}
\usepackage{booktabs}

\usepackage{filecontents}
\usepackage{xspace}
\usepackage{xcolor}
\usepackage{hhline}
\usepackage{multirow}
\usepackage{soul}

\usepackage{array}
\usepackage{comment}
\usepackage{listings}
\usepackage{dsfont}
\usepackage{algorithmic}
\usepackage{xurl}
\usepackage{setspace}
\usepackage[flushleft]{threeparttable}
\usepackage{enumerate}
\usepackage{float}
\usepackage{subcaption}
\usepackage{graphicx}
\usepackage{breqn}
\usepackage{enumitem}

\usepackage{caption}
\usepackage{ltablex}

\usepackage{ifluatex}

\usepackage{outlines}

\usepackage{comment}

\definecolor{darkkhaki}{rgb}{0.74, 0.72, 0.42}

\newcommand{\name}{\text{METIS}\xspace}

\newcommand{\maprerank}{\texttt{map\_rerank}\xspace}
\newcommand{\stuff}{\texttt{stuff}\xspace}
\newcommand{\mapreduce}{\texttt{map\_reduce}\xspace}
\newcommand{\chunksize}{\texttt{chunk\_size}\xspace}

\newcommand{\synthesismeth}{\texttt{synthesis\_method}\xspace}

\newcommand{\numchunks}{\texttt{num\_chunks}\xspace}
\newcommand{\interlen}{\texttt{intermediate\_length}\xspace}

\newcounter{packednmbr}
\newenvironment{packedenumerate}{\begin{list}{\thepackednmbr.}{\usecounter{packednmbr}\setlength{\itemsep}{0.5pt}\addtolength{\labelwidth}{-4pt}\setlength{\leftmargin}{2ex}\setlength{\listparindent}{\parindent}\setlength{\parsep}{1pt}\setlength{\topsep}{0pt}}}{\end{list}}
\newenvironment{packeditemize}{\begin{list}{$\bullet$}{\setlength{\itemsep}{0.5pt}\addtolength{\labelwidth}{-4pt}\setlength{\leftmargin}{2ex}\setlength{\listparindent}{\parindent}\setlength{\parsep}{1pt}\setlength{\topsep}{2pt}}}{\end{list}}

\newcommand{\tightcaption}[1]{\vspace{-0.4cm}\caption{{\normalfont{#1}}}\vspace{-0.3cm}}
\newcommand{\tightsection}[1]{\vspace{-0.2cm}\section{#1}\vspace{-0.2cm}}

\newcommand{\tightsubsection}[1]{\vspace{-0.25cm}\subsection{#1}\vspace{-0.2cm}}

\newcommand{\eg}{{\it e.g.,}\xspace}
\newcommand{\etc}{{\it etc.,}\xspace}
\newcommand{\ie}{{\it i.e.,}\xspace}

\newcommand{\mypara}[1]{\vspace{0.05cm}\noindent{\bf {#1}:}~}

\newcommand{\myparaq}[1]{\smallskip\noindent{\bf {#1}?}~}

\definecolor{backcolour}{rgb}{0.96,0.96,0.96}
\definecolor{codegray}{rgb}{0.5,0.5,0.5}
\definecolor{deepblue}{rgb}{0,0,0.6}
\definecolor{deepred}{rgb}{0.6,0,0}
\definecolor{deepgreen}{rgb}{0,0.5,0}
\lstdefinestyle{mystyle}{
    backgroundcolor=\color{backcolour},   
    commentstyle=\color{codegreen},
    morekeywords={self, True},
    keywordstyle=\color{deepblue},
    numberstyle=\tiny\color{codegray},
    emph={MyClass,__init__,EncodingType,Image},
    emphstyle=\color{deepred},
    stringstyle=\color{deepgreen},
    basicstyle=\ttfamily\footnotesize,
    breakatwhitespace=false,         
    breaklines=true,                 
    captionpos=b,                    
    keepspaces=true,                 
    numbers=left,                    
    numbersep=5pt,                  
    showspaces=false,                
    showstringspaces=false,
    showtabs=false,                  
    tabsize=1
}

\begin{document}

\newcommand*{\affmark}[1][*]{\textsuperscript{#1}}
\newcommand*{\affaddr}[1]{#1}


\title{\name: Fast Quality-Aware RAG Systems with Configuration Adaptation}


\author{Siddhant Ray}
\affiliation{
  \institution{University of Chicago}
  \city{}
  \country{}
}

\author{Rui Pan}
\affiliation{
  \institution{Princeton University}
  \city{}
  \country{}
}

\author{Zhuohan Gu}
\affiliation{
  \institution{University of Chicago}
  \city{}
  \country{}
}

\author{Kuntai Du}
\affiliation{
  \institution{University of Chicago / TensorMesh }
  \city{}
  \country{}
}

\author{Shaoting Feng}
\affiliation{
  \institution{University of Chicago}
  \city{}
  \country{}
}

\author{Ganesh Ananthanarayanan}
\affiliation{
  \institution{Microsoft}
  \city{}
  \country{}
}

\author{Ravi Netravali}
\affiliation{
  \institution{Princeton University}
  \city{}
  \country{}
}

\author{Junchen Jiang}
\affiliation{
  \institution{University of Chicago / TensorMesh }
  \city{}
  \country{}
}

\renewcommand{\shortauthors}{Ray et al.}


\begin{abstract}

RAG (Retrieval Augmented Generation) allows LLMs (large language models) to generate better responses with external knowledge, but using more external knowledge causes higher response delay.
Prior work focuses either on reducing the response delay (\eg better scheduling of RAG queries) or on maximizing quality (\eg tuning the RAG workflow), but they fall short in systematically balancing the {\em tradeoff} between the delay and quality of RAG responses. 
To balance both quality and response delay, this paper presents \name, the first RAG system that {\em jointly} schedules queries and adapts the key RAG configurations of each query, such as the number of retrieved text chunks and synthesis methods. Using four popular RAG-QA datasets, we show that compared to the state-of-the-art RAG optimization schemes, \name reduces the generation latency by $1.64-2.54\times$ without sacrificing generation quality.

\end{abstract}




\maketitle



\section{Introduction}
\label{sec:intro}


Retrieval-augmented generation (RAG) is a popular LLM inference technique that augments an LLM inference query with relevant text chunks, or ``context'', {\em retrieved} from a large corpus.\footnote{RAG vs. long-context models is an active field of research, with the industry widely deploying RAG for its task-focused model inference quality and better resource-sharing capabilities~\cite{rag_industry}.}
{\bf RAG systems}, which include retrieval and LLM inference,\footnote{Though RAG sometimes refers to the retrieval step in this work, a RAG system includes both retrieval and LLM inference based on the retrieved texts, and we aim to optimize the whole pipeline.}, have found many use cases in QA tasks, personal assistants, chatbots, and LLM-powered search~\cite{ouyang2024contextaware,10.1145/3626772.3657834}.
While RAG can enhance the quality (accuracy and relevance) of LLM-generated responses~\cite{balaguer2024ragvsfinetuningpipelines,mao-etal-2024-rag,zhu2024ragevalscenariospecificrag,zhao-etal-2024-optimizing,nguyen2024enhancingqadomain}, RAG queries are inherently slow as they need more compute and memory resources to process the long input context to answer a query~\cite{acl_rag_tutorial, 10.1145/3637528.3671445,leng2024longcontextragperformance}. 
Thus, it is essential to balance {\em high response quality} and {\em low response delays} in RAG inference systems.

Past research efforts have optimized RAG, regarding either response quality or response delay, but they fall short in optimizing the {\bf quality-delay tradeoffs} of RAG.
RAG queries have an associated {\em RAG configuration} which describes how and how much data to input for the query (more in \S\ref{sec:background})~\cite{wang-etal-2024-searching, simon2024methodologyevaluatingragsystems, xie2024weknowragadaptiveapproachretrievalaugmented}.
One line of prior work focuses on reducing response delay through better query scheduling (\eg GPU allocation and inference batching) for RAG queries~\cite{lin2024parrotefficientservingllmbased,shahout2024fastinferenceaugmentedlarge,tan2024teolaendtoendoptimizationllmbased,infercept, lin2025teleragefficientretrievalaugmentedgeneration}, without adapting the RAG configuration themselves. 
An alternate line of work focuses on maximizing generation quality by tuning the configurations of RAG queries~\cite{xie2024weknowragadaptiveapproachretrievalaugmented,jeong-etal-2024-adaptive,tang2024mbaragbanditapproachadaptive}, but this is often done at the cost of longer response delay. 

The RAG configuration {\em simultaneously} affects generation quality and response delay (\eg retrieving too many chunks for a simple RAG query may unnecessarily inflate delay without increasing quality). 
Unlike traditional data queries (\eg SQL) which specify the inputs and operators, RAG queries are inherently {\em under}-specified as they consist of a text query written in natural language~\cite{mombaerts2024metaknowledgeretrievalaugmented, guo2024lightragsimplefastretrievalaugmented, qian2024memoragmovingnextgenrag,jeong-etal-2024-adaptive} and do not directly specify the exact RAG configuration of its execution.

Moreover, {\em multiple} configuration knobs can influence the delay-quality tradeoffs.
For instance, besides how many chunks to retrieve, {\em how} to use them in the LLM's input involves two design choices---should the chunks be processed by the LLM jointly, or should the chunks be summarized first before being fed into the LLM together (and how long should a summary be).
Recent works also attempt to tune RAG configuration~\cite{jeong-etal-2024-adaptive,tang2024mbaragbanditapproachadaptive}, but they focus on either tuning individual knobs or maximizing quality at the cost of higher delay. However, tuning configurations across multiple knobs quickly hits a {\em prohibitive} combinatorial space (more in \S\ref{sec:motivation}) and requires optimizations to reduce the search cost.

What's more, the RAG configuration should be tuned {\em jointly} with scheduling. 
Consider two configurations: $A$ feeds all retrieved text chunks in one LLM input, and $B$ summarizes first each chunk with an LLM and then feeds the summaries to an LLM input for a final generation.
While $A$ (which calls the LLM once) is seemingly faster than $B$ (which calls the LLM multiple times), $A$ could be slower as it requires more GPU memory than $B$ and thus could be delayed in the scheduler queue.
Without making batching and configuration selection jointly, it would be difficult to avoid such pitfalls.

Finally, the impact of RAG configurations on quality-delay tradeoffs also varies {\em significantly} with queries. 
For example, to answer \emph{``In which country is the Kimbrough Memorial Stadium located?''}, the RAG may retrieve and analyze one text chunk about the stadium.
In contrast, to answer \emph{``Compare NVIDIA's operating cost over the first three quarters of 2024 and identify the highest one''}, the RAG may need multiple chunks, each containing the quarter's operating cost, and process these chunks jointly, instead of reading them separately. The above examples illustrate queries differ in {\em complexity} (more in \S\ref{sec:design}), leading to needing different configurations per-query for optimal quality-delay tradeoffs.
Empirically, we show that picking RAG configuration {\em per-query} achieves $12-15\%$ higher quality and $2.5-3\times$ lower delay than using any fixed configuration across all queries in a dataset (\S\ref{fig:adaptation_benefits}). 
Thus, RAG configurations should be adapted on a {\em per-query} basis.

Yet, existing RAG systems, which hand-pick a {\em static} configuration offline based on a few example queries~\cite{kim2024autoragautomatedframeworkoptimization, rag_hyperparameter_tuning,autorag_hp,yao2024cacheblendfastlargelanguage}, lose out on quality or response time.

This paper presents \name, the {\em first} RAG system that adapts multiple configuration knobs on a per-query basis and jointly makes configuration selections and scheduling decisions (\ie which LLM inference in a batch) to optimize the delay-quality tradeoffs for RAG.

As this would require solving a joint combinatorial problem for every query, which can be prohibitively expensive (\S\ref{sec:motivation}),
\name tackles the challenge with a two-step approach.

First, \name~{\em prunes} the massive configuration space for each received query to a smaller yet promising one that contains configurations that likely yield high-quality output for the given query. 
Specifically, \name uses a separate LLM to estimate the query's {\em profile}, including how many pieces of information are required to answer the query and whether joint reasoning is likely required across these pieces of information (more in \S\ref{ssec:llm_profiler_config}). 
The intuition of the query profiles is that they can effectively filter out undesirable RAG configurations.
For the earlier query example ``{\em Compare NVIDIA's operating cost over the first three quarters of 2024 and identify the highest one},'' the estimated profile would suggest that it involves at least three separate pieces of information, so the number of chunks (one of the configuration knobs) should be at least three. 
It should be noted that the LLM-based profiler is an extra overhead in \name, but fortunately, its input only contains the RAG query itself and the metadata of the RAG database, which are orders of magnitude shorter than the long contexts in RAG,
so the estimation can be relatively {\em fast}, about 1/10 of the delay of the execution of the RAG query.

Using the narrowed configuration space, \name reduces the RAG response delays by {\em jointly} deciding the per-query configuration and query scheduling based on available resources (\S\ref{ssec:design_scheduler}). 
The insight is that within the pruned configuration space, the scheduler can make optimal configuration decisions without exploring the original, large configuration space and the implications on quality.

In short, \name's two-level design {\em loosely} decouples the problem into (1) pruning configuration space to a smaller yet promising range of configurations, which focuses solely on keeping the accuracy high, and (2) jointly optimizing configuration (within the narrowed range) and scheduling to optimize response delay by choosing configurations which best-fit into the GPU memory.

\begin{figure}
    \centering
    \includegraphics[width=0.95\linewidth]{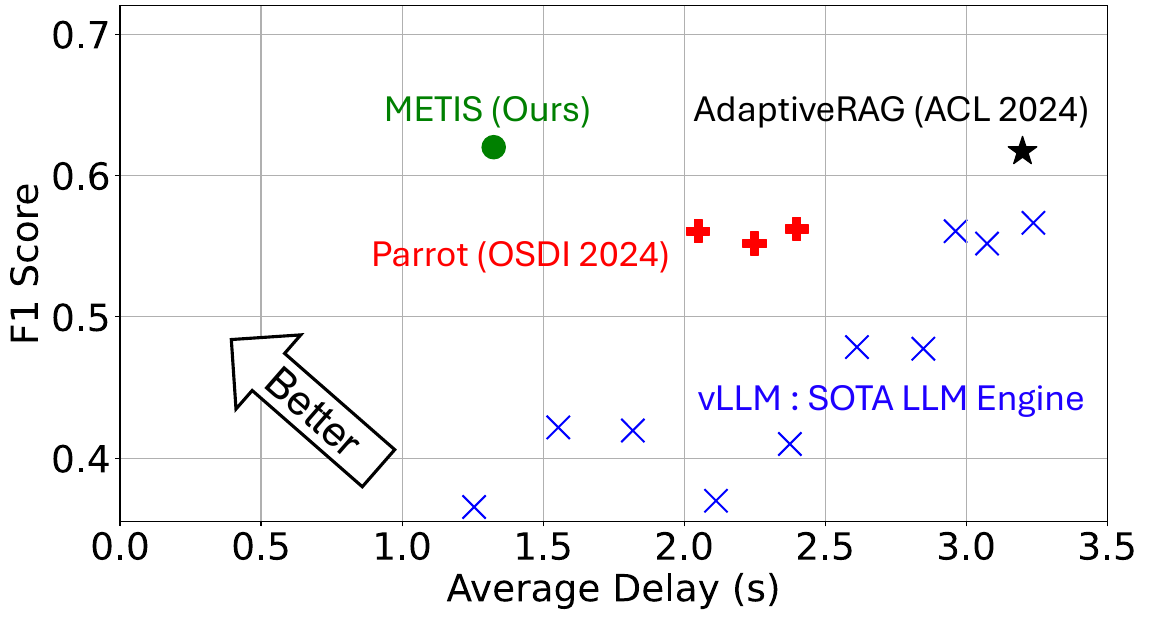}
    \caption{Performance of \name on the KG RAG FinSec~\cite{kgrag} dataset compared to the baselines. Full results shown in \S \ref{sec:eval}.}
    \vspace{-8pt}
    \label{fig:intro}
\end{figure}

We evaluate \name across four RAG datasets with diverse query profiles 
(\eg reasoning vs. domain-specific QA).
Figure \ref{fig:intro} shows a preview of our results.
Our key takeaways are as follows. 
When achieving the same or higher quality than the baselines, \name reduces the response delay by $1.6-2.8\times$ compared to the latest vLLM (a state-of-the-art serving engine), Parrot (the latest LLM query-scheduling method), as well as AdaptiveRAG (the latest RAG configuration-tuning method).
\name also achieves $1.8-4.5\times$ higher throughput compared to these baselines when achieving the same response delay and same/higher quality. 

The general concept of using LLMs to guide system tuning is not exactly new~\cite{298701, ong2025routellmlearningroutellms}, but our key contribution lies in applying the concept to RAG systems, through joint scheduling with resource-aware configuration selection, leading to significantly better resource sharing (\S\ref{ssec:mapping}, \S\ref{ssec:design_scheduler}). {\color{black} \name is the first work which (a) shows the importance of tuning multiple RAG knobs; (b) profiles multiple knobs and adapts them simultaneously; and (c) is the first LLM system to introduce resource-quality tradeoff in its RAG decisions.}

\vspace{-5pt}




\section{RAG systems and configurations}
\label{sec:background}

As an LLM often does not have domain-specific or up-to-date knowledge, LLM applications commonly employ RAG to supplement LLM inference with external knowledge to generate high-quality responses. Despite the growth of model context length, using RAG to pinpoint the relevant context is still significantly cheaper in terms of {\em resource cost} (GPU requirement), {\em latency}, and {\em memory consumption} (KV Cache size). For general-purpose QA pipelines, RAG is cost-efficient with retrieving targeted chunks based on semantic similarity to the query. Using LLMs with long-context documents in contrast has much higher GPU memory usage and delay~\cite{lin2025teleragefficientretrievalaugmentedgeneration, NEURIPS2024_a5d8aba2, li2024retrievalaugmentedgenerationlongcontext}.

Before processing queries, a RAG system organizes background documents by splitting them into chunks (each with a fixed number of tokens), embedding each chunk using models like Bert~\cite{devlin2019bertpretrainingdeepbidirectional,mteb_benchmark}, and storing the embeddings with the chunks in a vector database.

Processing a RAG query involves two main steps:
\begin{packeditemize}

\item {\em Retrieval:} The RAG system retrieves one or more relevant context chunks from the database by comparing the query's embedding, (using the same embedding model as for database indexing), with the stored embeddings.

\item {\em Synthesis:} After retrieving the relevant chunks, the RAG system combines these chunks and the RAG query to form a single/multiple LLM call(s) to generate the response. 

\end{packeditemize}
Retrieval is computationally lightweight and much faster than synthesis (> $100\times)$, so the response delay is typically dominated by the synthesis step~\cite{ralmspec}. 

\begin{figure}
    \centering
    \includegraphics[width=0.95\columnwidth]{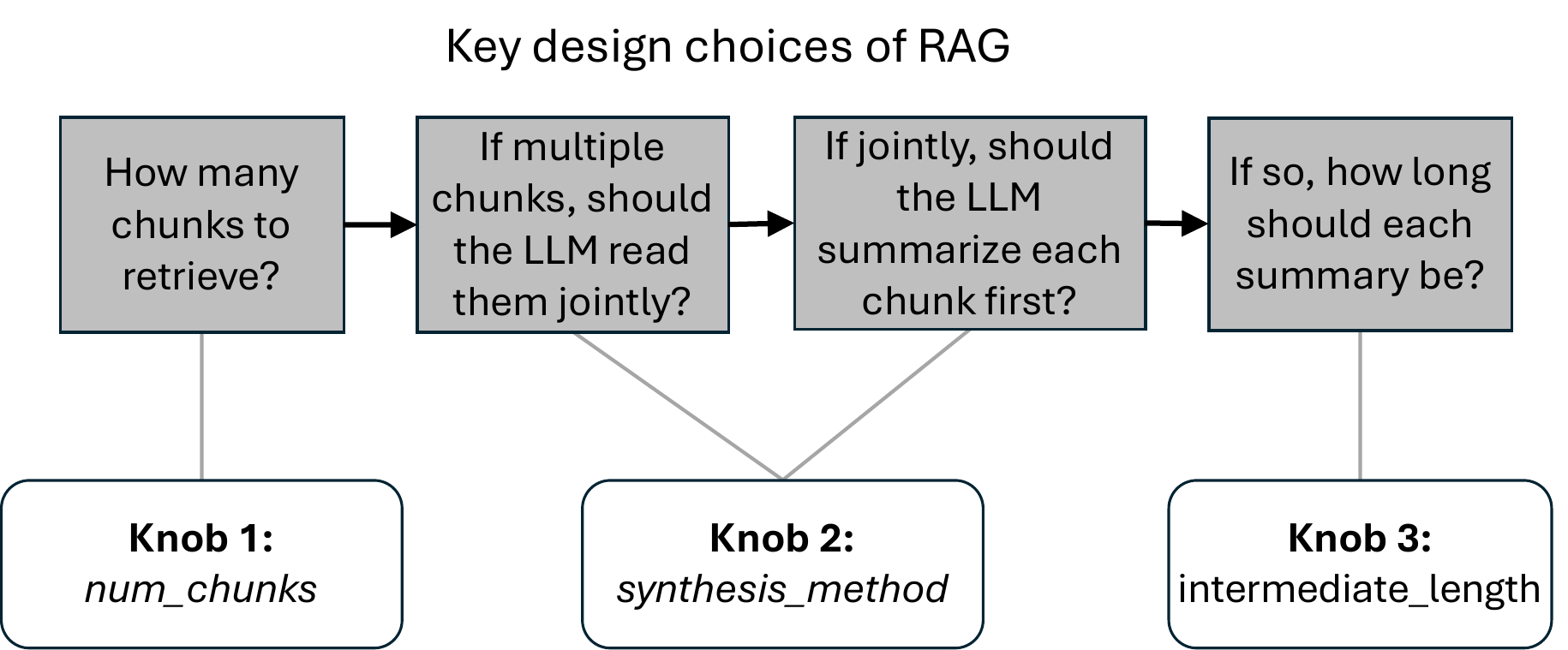}
    \tightcaption{The configuration knobs adapted by \name are derived from key design choices of RAG systems.
    }
    \label{fig:config_knobs}
\end{figure}

\mypara{RAG configuration}
This work focuses on optimizing three configuration {\bf knobs}, illustrated in Figure~\ref{fig:config_knobs}, which are derived from key design questions that affect RAG performance in terms of response delay and quality:  
\begin{packeditemize}
\item {\em How many chunks to retrieve} (\numchunks): The number of context chunks directly affects the delay of the synthesis step, with more computation needed to process the longer sequences with more chunks. 
In the meantime, retrieving too few chunks risks low response quality if the retrieved chunks do not contain enough useful information. 
\item {\em How to synthesize} (\synthesismeth):
If the LLM should read the chunks separately, RAG uses the LLM to generate one answer for the query using each chunk separately and picks the output with the highest confidence, which is called \maprerank.
This often incurs the least computation but can cause low quality if the useful information is scattered in different chunks, in which case the LLM should read the chunks jointly.
The RAG system can  feed these chunks in the LLM input directly by concatenating them within a single prompt (called \stuff) or to create a shorter summary for each chunk first before feeding the summaries and the query into the LLM to generate the final response (called \mapreduce).
\stuff needs less computation than \mapreduce, but risks degraded output quality for long inputs due to the lost-in-the-middle problem~\cite{lost_in_the_middle}. 
\item {\em How long is each summary} (\interlen):
Finally, if the LLM produces the summary for each chunk based on the user query, the length of each summary greatly affects the quality and response of \mapreduce---shorter summaries yield lower delay but also risk not feeding enough information to the final LLM inference. 
\end{packeditemize}

In this work, while we focus on universal RAG knobs which affect quality and delay common to {\em all} RAG systems, \name can be extended to other tunable knobs (\eg some RAG system may dynamically choose the embedding model, retrieval index or serving LLM). \name' design is extensible to any RAG configuration knob based on the query profile.

\mypara{Performance metrics}
We evaluate the performance of a RAG system using two metrics:
\begin{packeditemize}
\item \textit{Response quality} calculates the F1 score of the generated response against the ground truth. The F1 score is the harmonic mean of precision (\# correctly generated words) and recall (\# of correct words successfully generated). This metric is widely used in prior works~\cite{ru2024ragcheckerfinegrainedframeworkdiagnosing,simon2024methodologyevaluatingragsystems,10.1145/3626772.3657834}.
\item \textit{Response delay} measures the time elapsed from when the RAG system receives a RAG request to when it completes generating the response.  
\end{packeditemize}

\begin{figure}
    \centering
    \includegraphics[width=\linewidth]{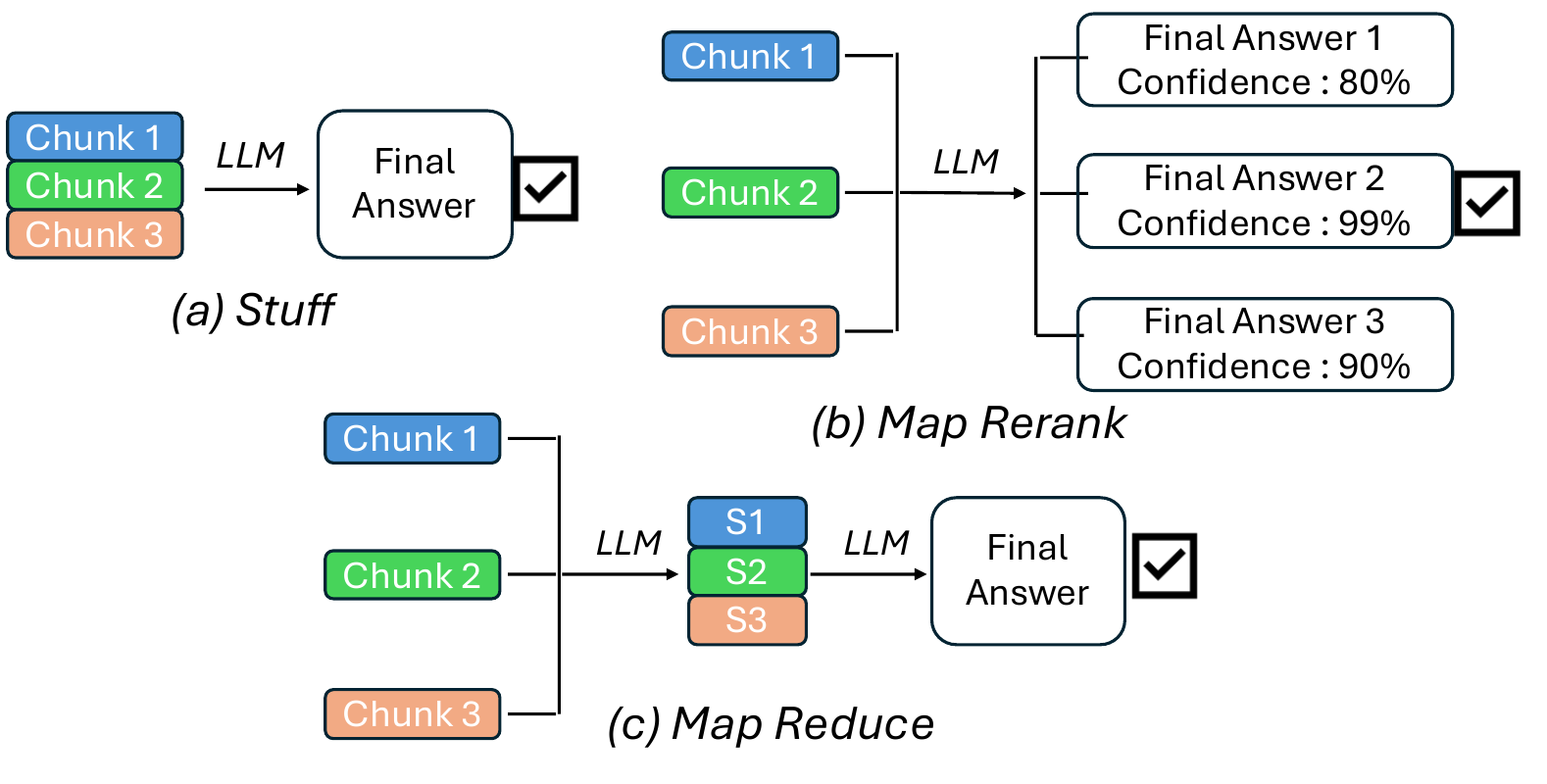}
    \tightcaption{Illustration of different RAG synthesis methods, which have various LLM reasoning capabilities.}
    \label{fig:synthesis_methods}
    \vspace{-2pt}
\end{figure}

Next, we show that these knobs need to be properly tuned on a per-query basis to achieve optimal tradeoff between quality and delay in \S\ref{sec:motivation}.


\section{Towards better quality-delay tradeoffs}
\label{sec:motivation}

\begin{figure*}[t]
    \centering
    \includegraphics[width=0.9\linewidth]{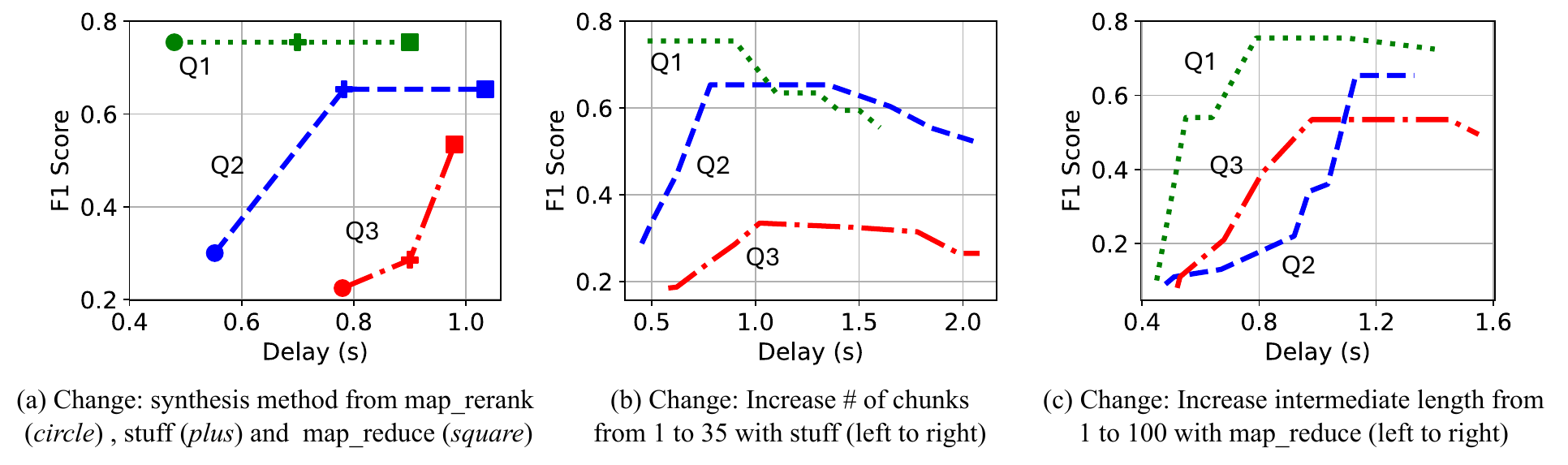}
    \vspace{-5pt}
    \caption{Varying each RAG configuration knob leads to different quality-latency tradeoffs, and these tradeoffs differ across queries (Q1 in green, Q2 in blue, and Q3 in red).}
    \vspace{-5pt}
    
    \label{fig:lat_acc_tradeoffs}
\end{figure*}



Prior work on RAG either optimizes for lower delay or higher quality, \ie
the first picks static configurations and focuses on reducing the delay by smart scheduling and resource allocation~\cite{lin2024parrotefficientservingllmbased,shahout2024fastinferenceaugmentedlarge,tan2024teolaendtoendoptimizationllmbased} and the second picks RAG configurations to maximize quality without regard to resource usage or delay~\cite{xie2024weknowragadaptiveapproachretrievalaugmented,jeong-etal-2024-adaptive,tang2024mbaragbanditapproachadaptive}. 
For the first time, we explore the potential of optimizing the {\bf \em quality-delay tradeoffs} for RAG.

To improve the delay-quality tradeoff, our insight is that 
quality and delay should jointly be optimized in this large tradeoff space created by the choice of RAG configuration knobs. 
Importantly, the configurations with better quality-delay tradeoffs {\em vary} significantly across queries. 

To showcase this observation, we use three queries from Musique~\cite{musique}, a popular reasoning QA dataset (\S\ref{ssec:eval_setup}). 
\begin{packeditemize}
\item {\bf Q1:} \emph{``In what county was William W. Blair's born?''}
\item {\bf Q2:} \emph{``Are Alison Skipper, Diane Gilliam Fisher, and Rachel McAdams from the same country?''}
\item {\bf Q3:} \emph{``When and why did the Voyager 1, the spacecraft that detected storms on Neptune, leave our solar system?''}
\end{packeditemize}
We chose queries with different natural language complexity and reasoning, Q1 being relatively less complex than Q2 and Q3. Then, we adjust the value of each configuration knob in order to quantify {\bf each knob's impact on the quality-delay tradeoffs} in each of the queries. 


\mypara{Impact of synthesis method}
Figure~\ref{fig:lat_acc_tradeoffs} (a) changes the synthesis method and shows its effect on the quality-delay tradeoff, while keeping the other RAG configuration knobs constant. We vary the synthesis method as \maprerank, \stuff, and \mapreduce from left to right. The insight is that the optimal synthesis method that strikes the best quality-delay tradeoff (closest to the top left corner) differs significantly across the different queries.

For simple queries like Q1 (green), quality plateaus for more complex synthesis methods (\stuff and \mapreduce). 
Because it only needs a single piece of context, \maprerank which processes chunks in isolation suffices, whereas cross-chunk reasoning (\stuff and \mapreduce) adds undue delay ($2\times$) without improving quality.

For queries such as Q2 (blue) that require {\em cross}-chunk reasoning, \stuff and \mapreduce provide significant quality improvements (35\% increase) by processing chunks jointly. 

For more complex queries, such as Q3  (red), which require even more reasoning and information (why Voyager 1 left has multiple reasons), methods like \mapreduce improve quality (30\% increase) by removing unnecessary text in the mapper phase, to help the LLM focus on the relevant content.

\mypara{Impact of the number of retrieved chunks}
Figure~\ref{fig:lat_acc_tradeoffs} (b) fixes the synthesis method (\stuff) and shows the impact of the number of retrieved chunks (1-35) on quality and delay. 

Simple queries, like Q1 (green), can often be answered using just one or two chunks (needs only birth county). For more complex queries, Q2 (blue) and Q3 (red), increasing the number of chunks (1-15) improves the likelihood of retrieving all relevant context and improves quality.

Blindly retrieving more chunks than necessary risks diluting the relevance of actual important information, due to commonly known problems such as ``lost-in-the-middle''~\cite{lost_in_the_middle, hsieh2024rulerwhatsrealcontext}.
In all three queries, retrieving more chunks beyond a point harms the quality (up to 20\% drop) and unnecessarily inflates delay (up to $3\times$). 
Hence we have a quality-delay tradeoff where increasing chunks up to a point helps quality but beyond that it increases delay while degrading quality.

\mypara{Impact of the intermediate output length}
Figure~\ref{fig:lat_acc_tradeoffs} (c) shows the impact of our third configuration knob, varying the intermediate output length (1-100) for \mapreduce synthesis methods on the quality-delay tradeoff. 
For simple queries like Q1 (green), short amounts of intermediate length are enough to answer the query (10-20 words). For more complex queries Q2 (blue) and Q3 (red), increasing the amount of intermediate length (70-100 words) provided helps the model with enough information to answer the query.

Overall, we see that RAG queries naturally vary in {\em complexity}, requiring differing levels of inter-chunk reasoning and varying numbers of context chunks. More complex queries, which require more reasoning and context, benefit from increased LLM computation, which can come at the cost of increased delay. Adding more context chunks helps to a point beyond which it harms the output quality and delay.

Thus, {\bf adapting RAG configuration on a {\em per-query} basis is crucial.} Figures \ref{fig:config_knobs}, \ref{fig:synthesis_methods}, \ref{fig:lat_acc_tradeoffs} {\em illustrate} tuning most popular RAG configuration knobs, however the tuning extends to more RAG configurations with richer tradeoff spaces (\S \ref{ssec:mapping}).

\begin{figure}[t]
    \centering
    \includegraphics[width=0.95\linewidth]{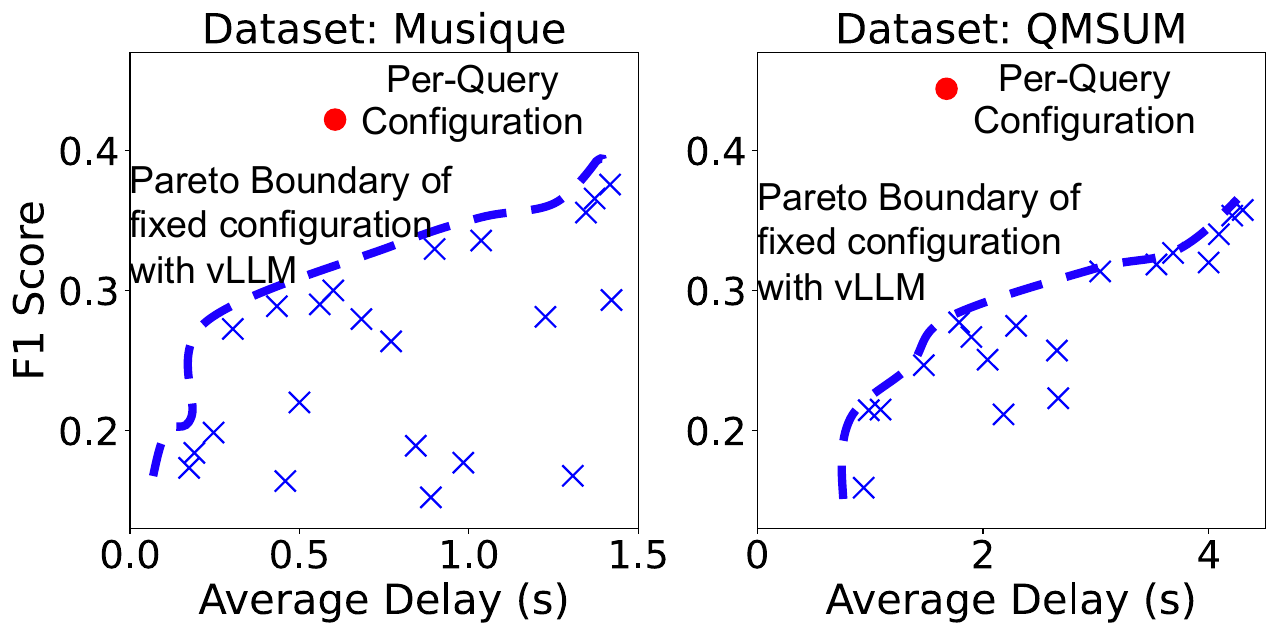}
    \caption{Per-query configuration can achieve significantly better quality-delay tradeoffs across queries compared to every fixed configuration choice.}
    \label{fig:adaptation_benefits}
    \vspace{-20pt}
\end{figure}

Figure~\ref{fig:adaptation_benefits} uses queries from two datasets (Musique and QMSUM, see \S\ref{ssec:eval_setup}) and shows that picking the best configuration for each query (the best configuration is the one with the lowest delay that achieves less than 2\% drop than the highest achievable quality)  achieves superior quality-delay tradeoff than picking any static configuration for all queries.
Choosing the configuration per-query allows up to 3$\times$ delay saving compared to static configurations which are the {\em closest} in quality.
Every single static configuration choice that achieves comparable delay has at least a 10\% quality drop. 


In spite of the potential benefits, per-query configuration adaptation faces challenges that hinder their real-world adoption. Each RAG query comes in plain text with practically no associated RAG configurations.
Moreover, the space of configurations grows {\em exponentially} with multiple knobs. 
For example, for a \mapreduce configuration, with 30 values for \numchunks and 50 values for \interlen leads to 1500 configurations for a query. Exhaustively profiling all configurations per-query and choosing the best is infeasible. 

Alternatively, if we profile periodically, we lose out on the potential configuration selection for {\em each} query, as variance in query profile leads to different quality-delay tradeoffs.
Profiling cost is also \emph{prohibitively} expensive as the LLM needs to be run with many synthesis methods, number of chunks \etc which require high GPU usage. Additionally, the delay of profiling can be $\sim$100$\times$ the inference delay due to multiple LLM calls during profiling. Online RAG queries have stringent requirements for GPU resource usage and end-to-end delay~\cite{shahout2024fastinferenceaugmentedlarge, tan2024teolaendtoendoptimizationllmbased}. This makes it hard to systematically decide what an optimal per-input configuration should be.

To truly achieve the benefit of per-query configuration adaptation, 
we need a \emph{smart} system to \emph{drastically} reduce to a useful configuration space, in a \emph{fast} and \emph{cheap} manner.

{\color{black} Finally, in the emerging space of agentic and reasoning RAG, profiling and configuration adaptation remains an integral part due the latency-sensitive nature of these systems. Optimally choosing the configurations, the focus of \name, remains crucial. We discuss this further in Section \ref{sec:discussion}.}


\section{\name: Enabling per-query configuration adaptation for RAG}
\label{sec:design}

\begin{figure}[t]
    \centering
    
    \includegraphics[width=0.95\linewidth]{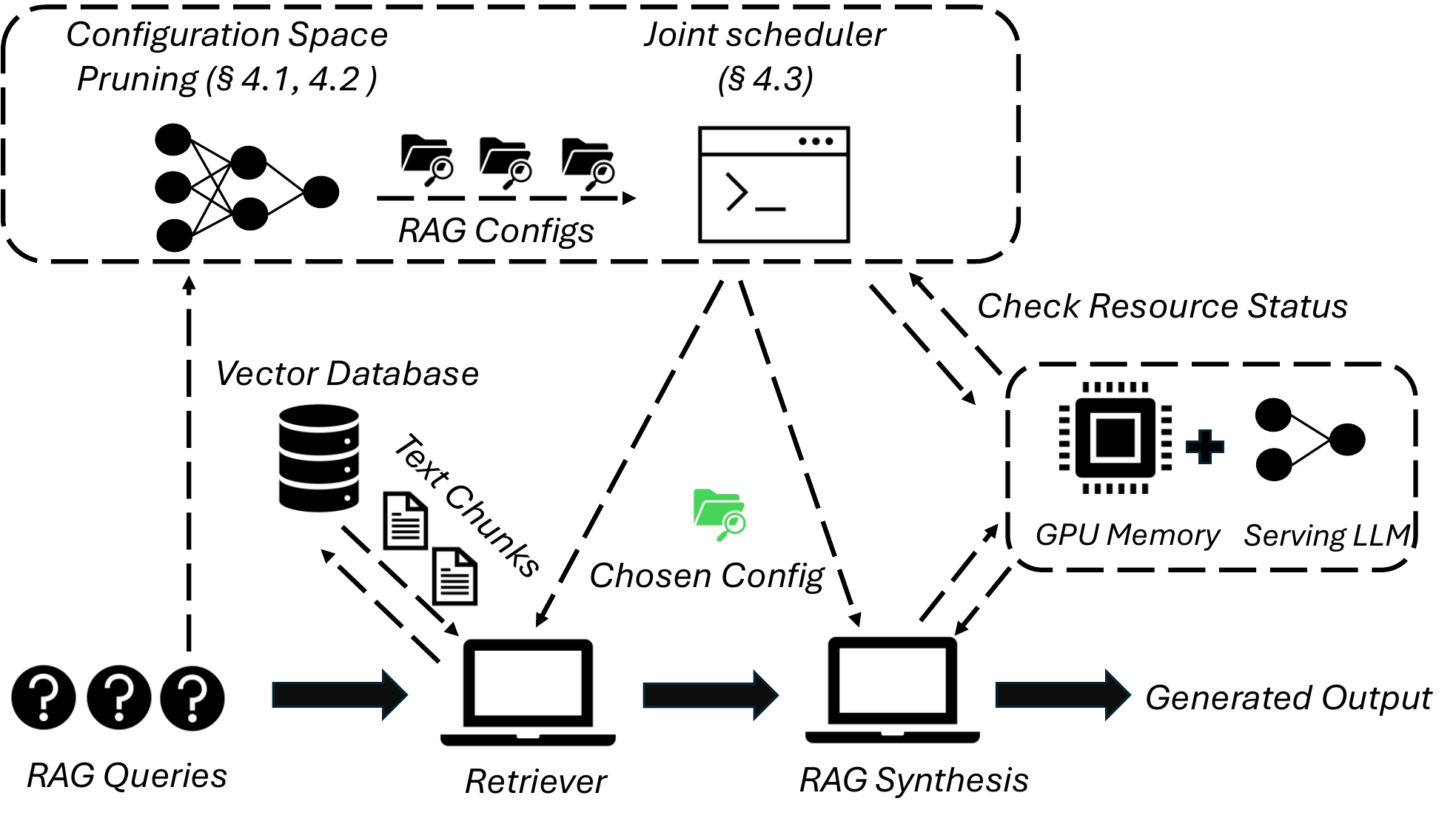}
    \tightcaption{\name consists of a RAG controller which performs configuration space pruning and joint scheduling.}
    \label{fig:design}
\end{figure}

We present \name, a novel system for serving RAG queries focusing on high generation quality and minimal delay. \name is a RAG controller (Figure~\ref{fig:design}) with two main components:

\begin{packeditemize}
    \item \emph{Pruning configuration space}: We estimate each query's profile (\S\ref{ssec:llm_profiler_config}) and reduce the RAG configuration space to a smaller yet promising one that still yields high generation quality (\S\ref{ssec:mapping}) (leading to a 50-100$\times$ reduction).
    \item \emph{RAG scheduler}: Within the pruned configuration space for the query, \name' scheduler chooses the best configuration for the query to achieve the best quality-latency trade-off based on the available system resources (\S\ref{ssec:design_scheduler}).
\end{packeditemize}


Once the configuration is chosen, the \name' executes the query using the chosen configuration---retrieving the selected number of chunks and uses the selected synthesis method to feed into the LLM's input.

%

\begin{figure*}[t]
    \centering
    
    \includegraphics[width=0.96\linewidth]{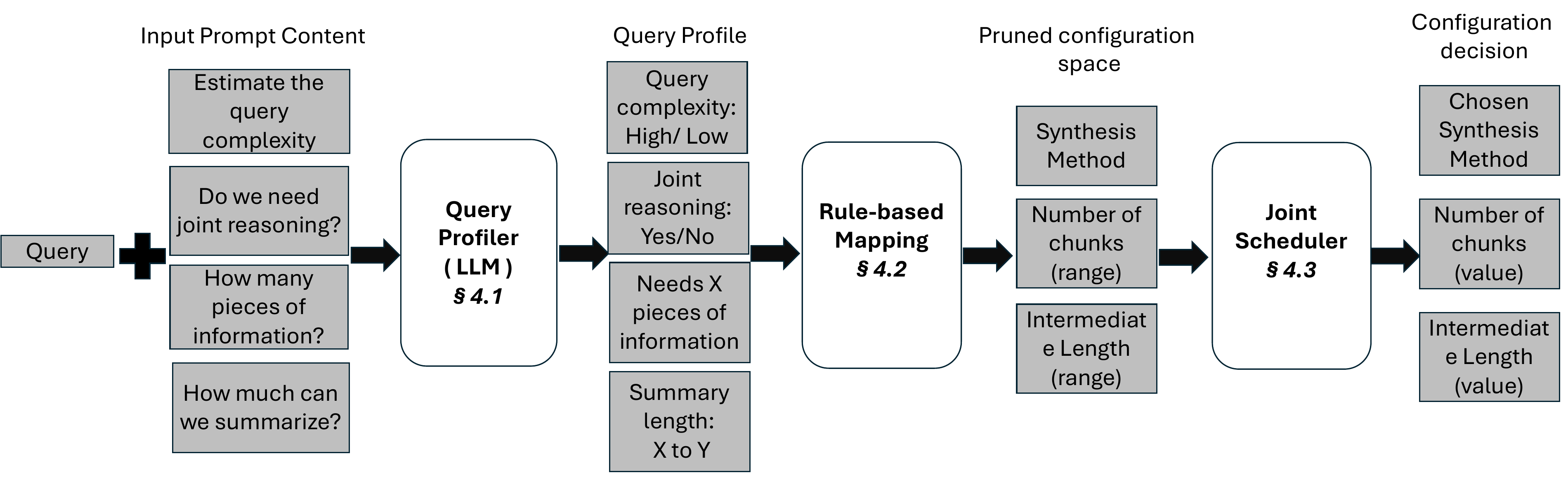}
    \caption{\name RAG configuration selection workflow.}
    \vspace{-5pt}
    \label{fig:workflow}
\end{figure*}

\subsection{Estimating a query's profile}
\label{ssec:llm_profiler_config}


\mypara{Query profile} To choose the correct RAG configurations, the first step of \name is to create the profile of the query (as we see in Figure \ref{fig:workflow}) by querying an LLM (we call this LLM \textit{query profiler}). We ask the query profiler to estimate four high-level dimensions for each query.

\begin{packeditemize}
\item {\em Query complexity} refers to the intricacy of the query itself. Queries with less complexity are more like simple yes/no questions, while queries with high complexity are more like why questions, which require deeper reasoning than yes/no questions.
As a result, it requires more LLM computation to correctly answer complex queries. The output for this dimension is binary ``High/Low''



\item {\em Joint reasoning requirement} describes whether multiple pieces of information are needed to answer the query.
Even relatively simple queries may require joint reasoning (\eg checking whether the annual income from two years is the same). The output for this dimension is binary ``Yes/No''

\item {\em Pieces of information required} refers to the distinct, standalone pieces of information required to fully answer the query (\eg the annual income from how many years is required to draw the trend of annual income). The output for this dimension is a number from 1-10.

\item {\em The length of the summarization:} If the query is complex and needs a lot of different information, it is often necessary to first summarize the relevant information chunks first (to reduce the noise inside these chunks) and then generate the final answer from these summaries. The output for this dimension is a number from 30-200.
\end{packeditemize}

\name is not the first to use query profile as a metric for deciding RAG configurations, it extends upon methods like AdaptiveRAG~\cite{jeong-etal-2024-adaptive} which have used LLM's to estimate query profile but they only focus on one dimension (the number of chunks to retrieve). {\color{black} However \name is the first LLM system to introduce resource-quality tradeoff in its RAG decisions with \emph{multiple} RAG configurations. We show this tradeoff in our experiments, along with the impact of each dimension on the overall improvement, in Section \ref{sec:eval}}.

\mypara{Why the query profile {\em could} be estimated}
Estimating the aforementioned query profile is feasible, not only because of the reasoning power of LLMs\footnote{We have tested both GPT and Llama models as the profile query-profiler, and they yield similarly impressive results (\S\ref{sec:eval}).} in analyzing natural language queries, but also because we provide sufficient information to the LLM-based profiler. 
\name feeds the profile estimator with not only the query, but also a \emph{metadata} of the database that contains the background document.

The metadata is a short description about the type of content in the database and its data size (\chunksize). 
Specifically, we use a single-line summaries already attached to the original source datasets as the metadata of the dataset. 
For example, the metadata for the KG RAG Finsec's database ~\cite{kgrag} contains quarterly financial reports and questions of Fortune 500 companies with a \chunksize of 1000. It describes the content topics of the chunks with information such as revenue growth indicators, product release information, sales \etc. When presented with a query on financials of such a company, the LLM can use the metadata to decide questions like how much to summarize/how much reasoning is required. 
We give details on the prompt and the intuition to generate metadata for new datasets in Appendix \S\ref{sec:appendix}.

{\color{black} The profiler uses an expressive model, as it only sees the input query and the dataset metadata without the whole context required for the RAG query. Based on the profiling and the available resources, a much smaller model works for inference as RAG uses the retrieved context, rather than model weights, to answer the question. The profiler, though a larger LLM, is cost-bounded, as the input is much smaller ($\sim$100-1000$\times$)~\cite{leng2024longcontextragperformance} compared to the retrieved context.} 


It is important to acknowledge that for highly under-specified queries, it is hard for any model (even human) to reasonably estimate the query's profile. 
For an example query ``Compare current US Stock Market trends,'' the query profile here does not provide enough information (\eg how many years should the trend be derived from).
To answer such highly under-specified queries, more information about the dataset will unlikely help.\footnote{Maybe some chat history from the same user will help, but that is beyond the scope of this work.}

Moreover, we observed that extra information does not significantly improve the profiler's estimates.
For instance, in theory, it helps to know the embedding algorithm  used by RAG. 
Yet, the embedding models perform similarly overall across queries and datasets under our consideration. 
This explains their limited contribution to the profiler, though more future work is needed to understand the wider implications.

\begin{algorithm}[t]
\SetKwInput{KwData}{Input}
\SetKwInput{KwResult}{Result}
\caption{Rule based mapping algorithm}\label{alg:1}
\KwData{\emph{Query complexity}, \emph{Joint reasoning required}}
\KwData{\emph{Pieces of information }, \emph{Summarization length range}}
\KwResult{synthesis\_method, num\_chunks, intermediate\_length}

  \eIf{Joint reasoning required == ``no''}{
    synthesis\_method = \maprerank
  }{
        \eIf{Query complexity == ``low''}{
            synthesis\_method = \stuff
        }
        {{
            synthesis\_method = \stuff, \mapreduce
        }
        }
    }
  num\_chunks =  [\emph{Pieces of information } , 3$\times$ \emph{Pieces of information}]

  intermediate\_length\_range = \emph{Summarization length range}
 
\end{algorithm}
\vspace{-8pt}

\subsection{Mapping query profiles to RAG configurations}
\label{ssec:mapping}

After \name obtains the query profile using the LLM, it performs rule-based mapping to generate values for RAG configuration knobs 
(\eg \synthesismeth etc. introduced in \S \ref{sec:background}). 
based on the query profiler's outputs.

\mypara{How we map and why the profile helps} To understand the role of query profiles, consider the following examples:
\begin{packeditemize}

\item \emph{``Who is the current CEO of NVIDIA?''} 
This query is not complex and does not require joint reasoning. Due to the query being simple with no reasoning required and one piece of information (name of CEO). 


\item \emph{``Which month had the highest NVIDIA's stock price the six months from January to June 2024?''} 
This query is simple but still needs to read information jointly, specifically six pieces of information (stock price for every month) 

\item \emph{``What are the reasons for NVIDIA's month-on-month stock price change from January to June 2024''}
This query is complex and needs to read multiple pieces of information jointly (stock prices, reasons for change etc.) As multiple reasons need to be analyzed here, summarizing all of the information first helps narrow down to relevant information and perform clearer reasoning (why the prices changed).

\end{packeditemize}

Algorithm \ref{alg:1} outlines the rule-based mapping process. This mapping is significantly helpful, it improves upon raw profiler outputs and converts them to usable RAG configurations. It reduces the cost of the profiler LLM by restricting it to provide short binary decisions only.

We decide the range of \synthesismeth selections based on two of the profile dimensions estimated in \S\ref{ssec:llm_profiler_config}, \ie the ``Query complexity'' and the ``Joint reasoning requirement''. Simple queries that don't need any reasoning can answered with \maprerank while queries that require joint reasoning need \stuff or \mapreduce.
We then decide the range of values for \numchunks based on the profile dimension of the ``Pieces of information required'', \ie $n$---specifically, we set the range of \numchunks to be $1-3$ times of $n$.
We do not directly set \numchunks at $n$, because it (1) gives some leeway for the retrieval logic (\eg typically Bert-embedding-based)\footnote{A typical RAG retriever will retrieve 2-3$\times$ more chunks than minimally required to provide sufficient information for the LLM inference~\cite{ragblog1,ragblog2}.} to find necessary information, and (2) provides the room for the scheduler to select the configuration that fits in available memory. 
Finally, we get the \interlen range from the ``summary length'' estimate, which is already a value range (derived from the query, metadata and chunk size). 

Algorithm \ref{alg:1} is {\em central} to \name' design to reduce to the space to our useful RAG configurations and this is {\em extendable} to other RAG configurations. For instance, a particular RAG pipeline might use an external re-ranker~\cite{Gao_2022, ma2023zeroshotlistwisedocumentreranking}, query re-writer~\cite{ma2023queryrewritingretrievalaugmentedlarge, jiang2023activeretrievalaugmentedgeneration} or perform an external web-search~\cite{singh2025agenticretrievalaugmentedgenerationsurvey} along with database retrieval. The mapping algorithm can map the profiling LLM's output (\eg of \emph{Query complexity}) and be used to guide such decisions for these newer RAG configurations.


Finally, it is important to note that the concept of \name belongs to an active research trend in the ML and systems community that leverages LLM outputs and mapping functions to guide real system decisions and optimizations, an example of which is \emph{LLM routing}~\cite{ong2024routellmlearningroutellms, hu2024routerbenchbenchmarkmultillmrouting, ding2024hybridllmcostefficientqualityaware, mohammadshahi2024routoolearningroutelarge}. While current LLM routers use trained LLMs to map decisions from query complexity to only choose from families of inference models (outside the realm of RAG), we differ by mapping the output to the configuration knob we run for RAG queries. 

Like these prior efforts, \name is a heuristic to best utilize the LLM-generated information to guide system optimizations. 
While it demonstrates remarkable improvement in practice, more work will be needed to complement it for better interpretability and robustness.

\subsection{Joint configuration-scheduling adaptation}
\label{ssec:design_scheduler}

Once provided with the narrowed range of each RAG configuration knob (\synthesismeth, \numchunks and \\
\interlen), we need to choose a RAG configuration, which is aware of the current system resource (GPU memory). If we pick configurations which do not fit in current memory, it will lead to additional queuing delay waiting for the GPU memory to free up.

We have \name's pruned configuration space where the quality is high, we now focus on choosing the best configuration which fits in memory, without focusing on quality.

\begin{figure}[t]
    \centering
    \scalebox{.99}{
    \includegraphics[width=.99\linewidth]{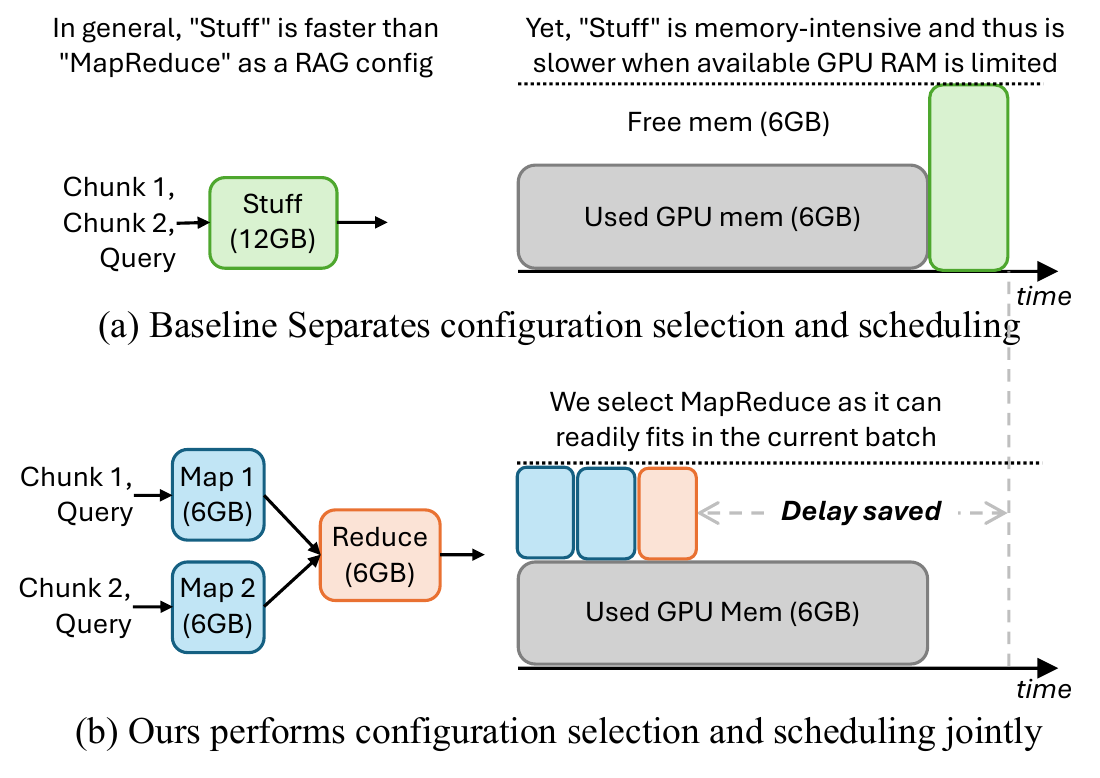}}
    \tightcaption{\name joint schedules RAG configurations with available GPU memory (chosen example - \mapreduce)}
    \label{fig:scheduling}
\end{figure}

\mypara{Why we need to choose the scheduling jointly}
We motivate the need for joint scheduling along with the RAG configuration choice in Figure \ref{fig:scheduling}. 

Consider a setup where we tune only one RAG configuration knob of \synthesismeth. Other knobs \numchunks and \interlen are fixed at 20 and 100 respectively. Let's assume both \stuff and \mapreduce are present in the pruned space. For the scheduling knob, we consider the amount of GPU memory available for the current batch.

Consider a baseline system which separates the joint decision from the scheduling and picks only the RAG configuration knob (\synthesismeth). It chooses the \stuff configuration knob as it has lower compute requirement, so given enough memory it should be fast. 

The baseline system in Figure \ref{fig:scheduling} (a) does not consider other jobs in the system and does not evaluate the amount of available resource to make its scheduling decision. Due to its long input length with 20 chunks, \stuff turns out to be memory-intensive. If the available GPU memory is low, \stuff doesn't fit in memory and needs to be queued. This ends up with \stuff being slow.

Jointly considering the available GPU memory with choosing the RAG configuration knob avoids this pitfall. For example, in Figure \ref{fig:scheduling} (b), if the original configuration was \stuff, \name can choose to use \mapreduce (based on the current GPU memory available). 

By doing so, \name can start putting the mappers which fit in memory, into the current \texttt{running_batch} of requests which fits in the GPU. While \mapreduce requires more compute, in this case, it benefits from being able to start execution much faster, as some of the mappers fit in memory.

\name does not need to wait for the GPU memory to free up and changes the configuration aware of system resource, to save delay and achieve a better quality-delay tradeoff.

\mypara{Jointly choosing the configuration knobs}
\name first provides us with a pruned range of configurations. A \emph{straw-man} solution is to pick a constant value from the across queries. (\eg the median value of the \numchunks ). While this is better than using one static configuration for all queries, it is still sub-optimal as it does not look at the current system resource availability. This prevents us from exploiting the best quality-delay tradeoff across RAG queries.

We use a \emph{best-fit} algorithm on underlying vLLM’s continuous batching to allow for variation in configurations across queries. \name first computes the GPU memory requirement for the RAG query from the RAG configuration knobs for every configuration in the pruned space. 
{\color{black}For RAG queries, the memory required (e.g., the KVCache size) is measured from the input token length, parameters of the serving model and the quantization (bytes per token). We keep a small 2\% buffer size on top of the measurement to deal with potential OOM crashes.} We measure the current \emph{available memory} on the GPU to see what can fit into the current batch. 

We then pick the {\em best configuration} from the pruned space that fits into the GPU. \name defines the best configuration as the one with overall highest memory requirement, from all which fit in memory. The insight here is that within the reduced range of good quality configurations, higher memory configurations correspond to expensive configurations (e.g. more number of chunks, higher intermediate length). In general, these configurations should lead to \emph{slightly higher quality} in the reduced space. For example, if the pruned space says \numchunks is 5-10 and the \synthesismeth is stuff and both 5 or 6 chunks can fit in memory, we choose 6 chunks. We don't pick a configuration that doesn't fit in GPU, so we would never choose more than 6 chunks. If we do that, the system will \emph{queue} the request inflating the delay.

After choosing the configuration that fits into the current \texttt{running_batch}, the vLLM engine is optimized to perform \emph{chunked_prefill}. However, even with \emph{chunked_prefill}, it can only offload parts of long prefill of \stuff requests which do not fit in the current batch and still inflates the queuing delay. Jointly scheduling RAG configurations enables efficient resource usage, which cannot be obtained by only relying on the output of the LLM profiler.

\vspace{-5pt}
\myparaq{What if none of the configurations fit in the GPU}
A main insight for \name's design comes from the observation that in general, the RAG-specific focused configurations can be \emph{loosely-decoupled} from the scheduling-specific configurations. \name tries to fit the best possible configurations into GPU memory after it gets the profiler's reduced configuration space. It can sometimes happen that the current GPU memory availability is too low and none of the profiler's configurations fit in the currently available GPU.

{\color{black} \name handles this is by falling back to a cheaper fixed configuration and ignoring the output space of the pruned configurations. As \name already have access to the query complexity profile, we can pick from cheaper configurations, which would meet the requirement for the current query. 

If the query does not require joint reasoning, we pick a \texttt{map\_rerank} with as many chunks that fit into available GPU memory. If joint reasoning is required, we pick a \texttt{stuff} with as many chunks that fit into memory. \name does not queue a configuration from outside the pruned range if none fit, but falls back to a fitting configuration just outside the range.}

This allows \emph{loose-decoupling} of the RAG configurations into a smaller space and then choosing configurations based on system resource availability. This also allows SLO-based constraints on RAG queries if certain queries have strict budgets on their generation latency.

\section{Refinements to \name}
\label{sec:refine}

In spite of it all, it is possible for the profiler to (sometimes) fail and in such cases, it is important to detect if \name's profiler fails on a query in a fast manner to prevent it from leading to bad RAG configurations. Also it is useful to decide how to provide feedback to \name to improve.

\begin{figure}[t]
    \centering
    \includegraphics[width=0.99\linewidth]{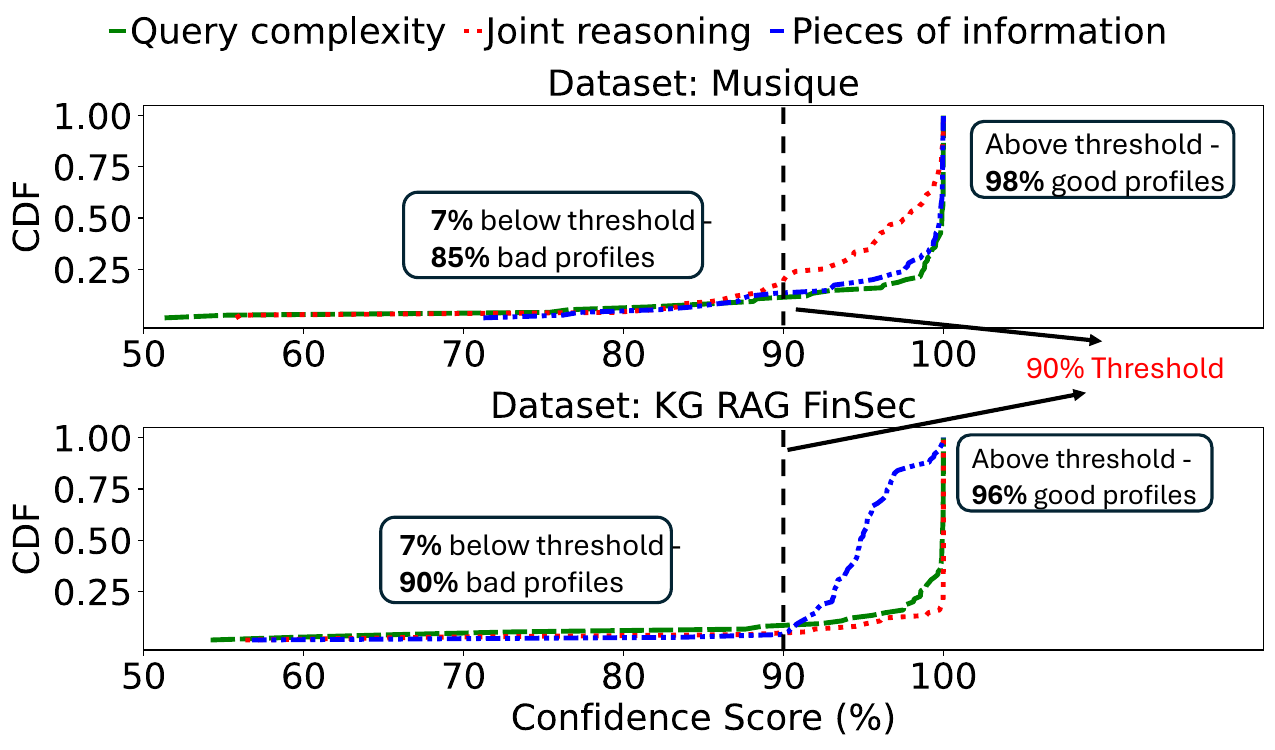}
    \tightcaption{Confidence score threshold for different profiler outputs is used to decide when not to use the profiler output.}
    \vspace{-2pt}
    \label{fig:logprob}
\end{figure}


\myparaq{When is the quality profile reliable}
\name uses LLM to generate the quality profile. Inspired by recent work in use of model confidence~\cite{xiong2024can,geng-etal-2024-survey,10.1145/3637528.3671470} as a quality metric, we use confidence scores for \name's LLM profiler as to measure the reliability of the profile provided. 
We obtain the confidence scores from the LLM's \emph{log-probs} values on the output (the logarithm of the confidence score, which is directly provided with the output with no extra overhead). 

We then threshold the confidence score using a confidence score threshold ($90\%$ across different datasets) to predict whether the quality profile derived from the quality profiler LLM is actually good (defined as whether the profile can lead to $10\%$ increase in F1-score or $1.5-2\times$ reduction in delay or both) or not.
Such $90\%$ threshold can be tuned for better performance, and we leave it to future work.
From Figure~\ref{fig:logprob}, we draw two conclusions. First, {\em over 93\% }of the quality profiles derived from LLM are of high confidence (\ie over 90\%). Further, for those high-confidence profile, over 96\% of them are good profiles, meaning that they can be used to improve quality, or reduce latency, or both.

To handle those cases where the quality profile is of confidence score {\em lower than 90\% }, \name will fall back to the pruned configuration space of recent 10 queries.





\myparaq{How to improve the profiler over time}
\name improves the query profiler LLM by profiling extra feedback prompt to this LLM.
We generate this feedback prompt by generating the most accurate output, which is obtained by performing inference on the most resource-demanding configuration (the \mapreduce configuration with a large number of input chunks (30) and a high value of intermediate length (300 tokens)) and then ask the quality profiler LLM what configuration it should choose based on the query \textit{and} the most {\em accurate} answer to that query.

The key insight is that, the most accurate answer to the query provides the quality profiler LLM \textit{extra knowledge} and thus can be used to further improve its decision.

To control the cost of generating feedback prompts, \name only generates the feedback prompt once every 30 queries and we only keep the \textit{last four} feedback prompts.






\mypara{The cost of \name' LLM quality profiler}
For the profiler LLM, we use a larger LLM as compared to the serving LLM (7B parameters). Using this has minimal cost, as \name \textit{only runs it on the query} itself and in \name as the query is at least 100$\times$ shorter than the context. Using this approach,  \name still saves cost as opposed to using a large LLM for inference (as shown in Section~\ref{sec:eval}). We also show that \name can use different closed and open-source LLMs as the profiler LLM for pruning and can still provide impressive delay reduction without hurting the accuracy in Section~\ref{sec:eval}.





\section{Implementation}
\label{sec:impl}

We implement \name in about 2K lines of code in Python on top of the state-of-the-art popular LLM serving engine vLLM~\cite{kwon2023efficient}. For the profiler used for configuration space pruning, we define a class \texttt{LLMProfiler} inheriting OpenAI's Chat Completion API~\cite{OpenAIAPI} interface (to invoke GPT-4o) and HuggingaceAPI~\cite{Wolf_Transformers_State-of-the-Art_Natural_2020} (to invoke LLama-3.1-70B) as models to profile the queries.

We use \texttt{Cohere-embed-v3.0}~\cite{cohere_embed} as a state-of-the-art embedding method. We construct a FAISS~\cite{douze2024faiss} index using the \texttt{IndexFlatL2} interface and perform L2-distance similarity search with \texttt{index.search(query_embedding, top_k)} on the chunk embeddings to retrieve for RAG inference.  We use the \texttt{LLMChain} interface from Langchain~\cite{Chase_LangChain_2022} in order to build efficient implementations of multiple synthesis methods. 

Finally, we use PyTorch's~\cite{Ansel_PyTorch_2_Faster_2024} library modules support to perform query-level memory profiling and measurement to implement the best-fit scheduling logic and request batching. Particularly, we use {\em pynvml} to construct \texttt{get_free_memory()} with its interfaces of \texttt{nvmlDeviceGetHandleByIndex} and \texttt{nvmlDeviceGetMemoryInfo} to measure the amount of GPU memory available. We measure the current \texttt{num-seqs} and \texttt{num-batched-tokens} within vLLM to calculate which configuration can be fit into the current batch, based on the GPU availability and the request's memory requirement.

\section{Evaluation}
\label{sec:eval}

The key takeaways from the evaluation are

\begin{packeditemize}
    \item \emph{Lower delay} : Across 4 task representative datasets for RAG QA, \name achieves $1.64 - 2.54 \times$ lower response delay compared to fixed configurations of comparable quality.
    \item \emph{Higher throughput} : \name achieves  $1.8 - 4.5\times$ higher throughput than RAG serving systems which use fixed configurations reaching similar quality.
    \item \emph{Negligible overhead} : \name' profiler's delay is negligible compared to the overall delay of the LLM's RAG inference.
\end{packeditemize}

\begin{figure*}[t]
\centering
     \includegraphics[width=.98\linewidth]{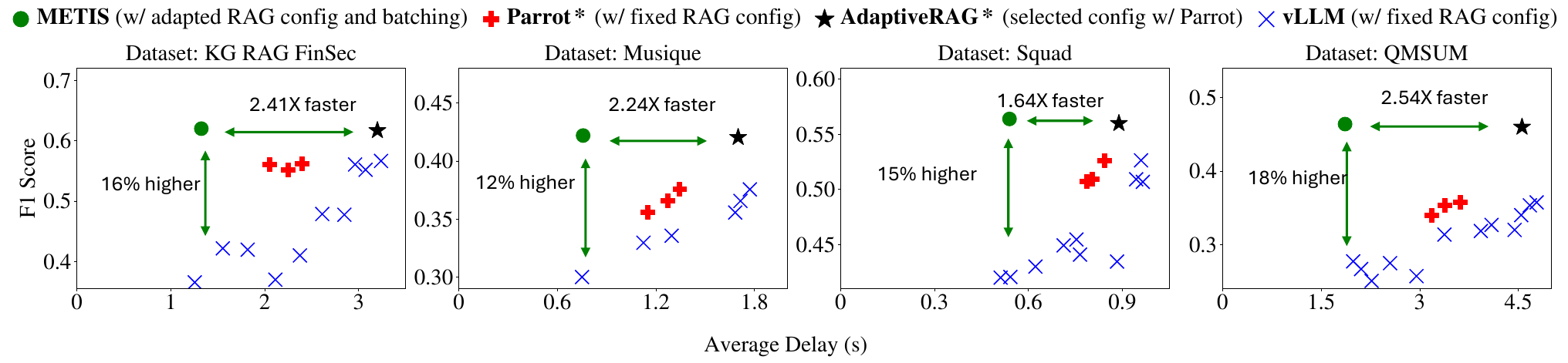}    
     \vspace{-0.2cm}
     \caption{\name achieves $1.64- 2.54\times$ lower delay compared to both best fixed configuration baselines and quality-optimized RAG configuration without sacrificing generation quality.} 
    \label{fig:e2e_gpt_lat}
    \vspace{-5pt}
\end{figure*}

\begin{figure*}[t]
\centering
     \includegraphics[width=.98\linewidth]{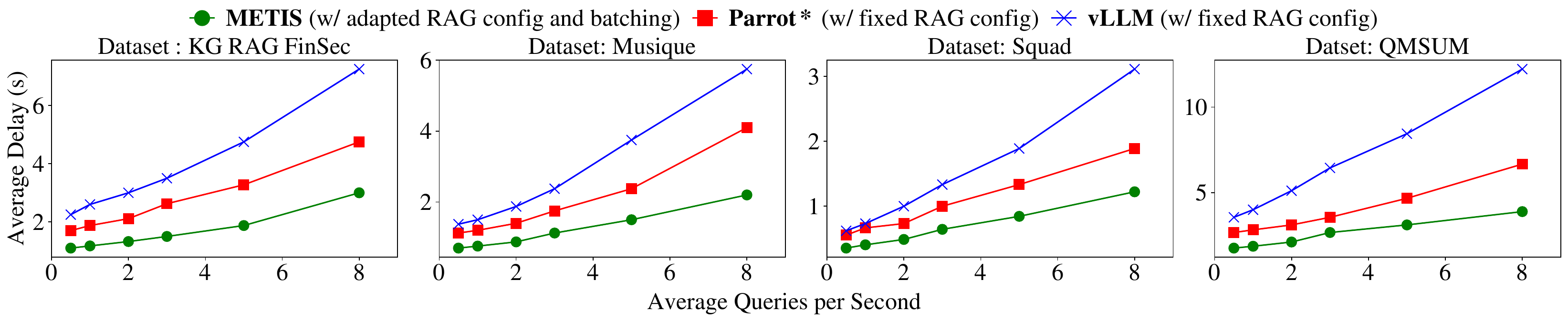}
     \vspace{-0.2cm}
    \caption{\name achieves $1.8 - 4.5\times$ higher throughput (at 1.8 seconds) than baselines which use fixed configurations of closest (not higher) quality.} 
    \vspace{-5pt}
    \label{fig:tpt}
\end{figure*}

\subsection{Setup}
\label{ssec:eval_setup}

\mypara{Models and hardware}: {\color{black}In RAG, models use external context to answer queries instead of the trained weights (model's embedded knowledge). The model extracts data from the external context and has stringent serving latency requirements. Hence RAG applications use smaller, instruction-tuned models. We evaluate \name on a popular models for RAG LLM inference, specifically the fine-tuned version of Mistral-7B-v3 and Llama3.1-70B for additional experiments as they are commonly used in all RAG QA workload serving. All models are fine-tuned such that they can take long contexts (up to 32K and 128K respectively). We apply AWQ-model quantization to both models.}

We use an NVIDIA A40 GPU server with
2 GPUs to benchmark our results. The server is equipped with
384GB of memory and two Intel(R) Xeon(R) Gold 6130 CPUs with
Hyper-threading and Turbo Boost enabled by default. We use 1 GPU to serve Mistral-7B-v3 and 2 GPUs to serve Llama3.1-70B.

\mypara{Datasets} We use multiple RAG QA datasets with various query profiles, in order to have task-representative workloads. Table \ref{tab:table_input} summarizes their input-output statistics.
\begin{packeditemize}
    \item Squad~\cite{squad}: Squad is a single-hop reading comprehension dataset, consisting of questions on articles, where the answer to every question is a segment from the corresponding reading passage.
    \item Musique~\cite{musique}: Musique is a multihop QA dataset with reasoning-based questions. It is designated to test LLM’s reasoning ability where one reasoning step critically relies on information from another. 
    \item KG RAG FinSec~\cite{kgrag}: KG RAG Finsec is part of a Knowledge Graph family of RAG datasets and focuses on financial domain questions from Fortune 500 companies. This dataset contains quarterly financial reports and queries need to read information for multiple chunks for answering.
    \item QMSUM~\cite{qmsum}: QMSUM is a human-annotated query-based multi-domain meeting summarization benchmark designed to test LLM's reasoning-based summarization capabilities. This dataset contains multiple meeting transcripts and queries to summarize relevant spans of meetings.
\end{packeditemize}

We build a retrieval database database by splitting the queries' contexts into fixed-sized chunks using Langchain~\cite{Chase_LangChain_2022} for the database, with \texttt{Cohere embed-v3.0}~\cite{cohere_embed} embeddings and FAISS~\cite{douze2024faiss} L2-distance similarity search in order to retrieve relevant chunks for RAG inference. To simulate a real RAG workload, {\color{black} we choose 200 queries from each dataset, and send them concurrently to \name using a Poisson distribution with an average arrival rate of 2 per dataset} . We report the results per dataset.

\begin{table}[t]
    \centering
    \scalebox{0.9}{
    \begin{tabular}{c  c  c  c}
        \toprule
          Dataset & Task Type & Input & Output  \\
         \cmidrule{1-4}
         Squad & Single hop QA & 0.4K - 2K & 5-10 \\
         Musique & Multihop QA & 1K - 5K & 5-20 \\
         KG RAG FinSec & Doc Level QA & 4K - 10K & 20-40 \\
         QMSUM & Summarization QA & 4K - 12K & 20-60 \\
         \bottomrule
    \end{tabular}
    }
    \caption{Input and output length (\# of tokens) distributions of the RAG datasets used in our evaluation.}
    \label{tab:table_input}
    \vspace{-20pt}
\end{table}

\mypara{Quality Metric} We adopt the following standard metric
to measure the generation quality.

\begin{itemize}
    \item F1-score is used to evaluate the \name's serving model's generated response (defined in \S \ref{sec:background}) It is the most widely adopted metric for evaluating RAG QA tasks~\cite{ru2024ragcheckerfinegrainedframeworkdiagnosing,simon2024methodologyevaluatingragsystems,10.1145/3626772.3657834}
\end{itemize}

\mypara{System Metrics} We adopt the following system metrics:

\begin{itemize}
    \item \emph{Delay} is used to measure the generation response delay of the model for every RAG query. We choose this system metric similar to other RAG serving papers~\cite{lin2024parrotefficientservingllmbased, tan2024teolaendtoendoptimizationllmbased, shahout2024fastinferenceaugmentedlarge}
    \item \emph{Dollar Cost} is used to measure the lower cost of using \name's profiler as compared to using larger serving models with fixed configurations having the closest accuracy.
\end{itemize}

\begin{figure}[t]
    \centering
    \includegraphics[width=0.9\linewidth]{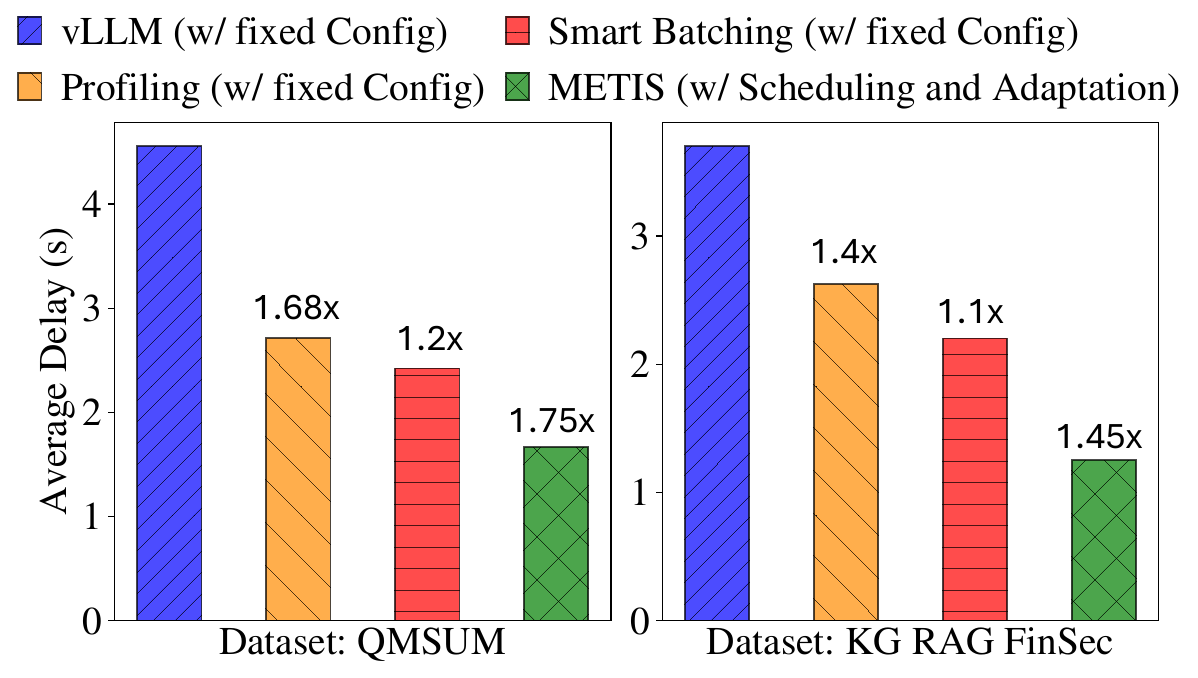}
    \caption{Understanding the delay improvement in \name}
    \label{fig:delay_red}
\end{figure}

\mypara{Baselines} We compare \name with the following baselines.
\begin{itemize}
    \item \emph{vLLM}: We serve RAG with vLLM with multiple static configurations across different queries.
    \item \emph{Parrot*}: We implement Parrot's~\cite{lin2024parrotefficientservingllmbased} configuration-based batching. Parrot* does not adapt the configuration per query. We compare with Parrot* using fixed RAG configurations which achieve the closest quality to us.
    \item \emph{AdaptiveRAG*}: We implement AdaptiveRAG's~\cite{jeong-etal-2024-adaptive}, query complexity-based RAG-configuration selection and choose the configuration which maximizes the F1-score, without considering the system resource cost.
\end{itemize}
\subsection{Overall improvement}

\mypara{Lower delay without sacrificing generation quality} Figure \ref{fig:e2e_gpt_lat} shows \name achieves delay reduction $1.64- 2.54\times$ over \emph{AdaptiveRAG*} with  no reduction in F1-score.  Over using fixed configurations of similar delay, served with both \emph{Parrot*} and \emph{vLLM}, \name achieves $12-18\%$ higher F1-score.

\mypara{Higher throughput at lower delay} Figure \ref{fig:tpt} shows \name achieves higher throughput compared to fixed configuration baselines when they choose the fixed-config which achieves the closest quality. {\color{black} \name has dynamic RAG configurations while vLLM and Parrot only allow a fixed configuration across queries. Baselines choose the configuration which achieves the highest average F1-score among all the fixed configurations but due to the static nature, every configuration achieves a lower F1-score compared to \name} Compared to  \emph{Parrot*} and \emph{vLLM}, \name achieves $1.8 - 4.5\times$ times higher throughput.

\mypara{Understanding \name' improvement}
\name's gains come from jointly selecting the configuration based on the available resource, along with performing scheduling. \name achieves higher quality than the fixed-config baselines as it is adapts the RAG-configuration per query. It reduces delay by resource-aware scheduling, making it better than fixed configurations which achieve closest quality.

\name achieves higher throughput as it is able to adapt configurations based on resource availability as compared to the baselines. Both \emph{Parrot*} and \emph{vLLM} schedule fixed RAG-configurations and cannot benefit from delay achieved by adapting the configuration like \name. \emph{Parrot*} can improve the delay over using fixed configurations with vLLM by $1.4-1.8\times$ but cannot improve the quality. {\color{black} Compared to  AdaptiveRAG*, \name achieves lower latency, as AdaptiveRAG* inflates serving latency without considering the cost of profiling or the cost of the configuration. AdaptiveRAG* also does not provide an interface to extend to multiple knobs.}

\subsection{Analyzing the gains from \name}

\mypara{Delay saving} Figure \ref{fig:delay_red} shows the contribution of every component of \name. We compare with vLLM's fixed configuration, which achieves the highest quality (blue bar). Using the profiler's outputs and choosing the median value every time (orange bar), we achieve $1.4-1.68\times$ reduction in delay. Next, we see the effect of batching (like Parrot*), by choosing the median value configuration and batching, we achieve $1.1-1.2\times$ reduction in delay. Finally, \name achieves even greater delay reduction by $1.45-1.75\times$ by adapting the configuration based on available GPU memory with batching.

\begin{figure}[t]
\centering
     \includegraphics[width=0.92\linewidth]{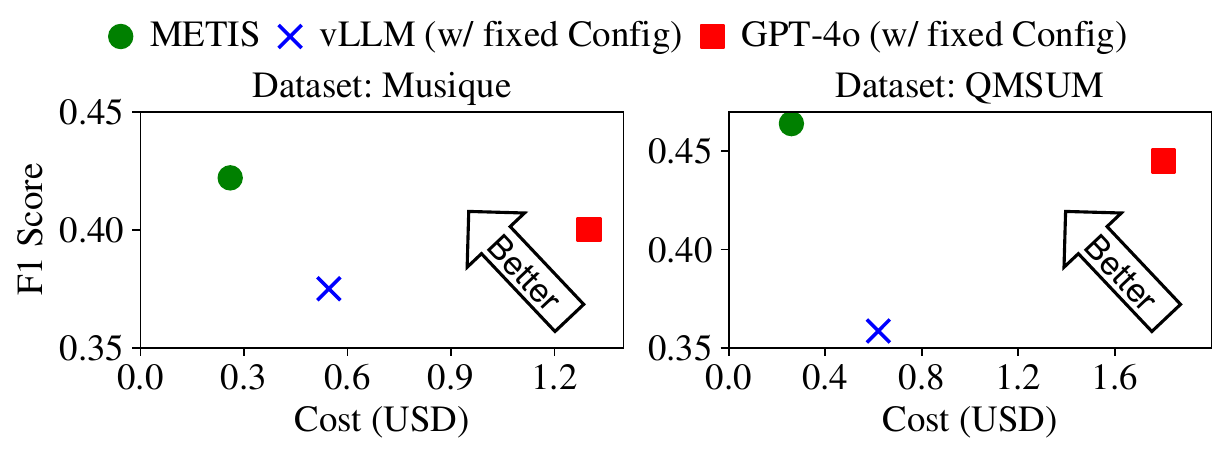}
    \caption{Even with increasing the inference model size, fixed configurations have $2.38-6.8\times$ higher cost and lower quality compared to \name.} 
    \vspace{-5pt}
    \label{fig:cost}
\end{figure}

\begin{figure}[t]
\centering
     \includegraphics[width=0.95\linewidth]{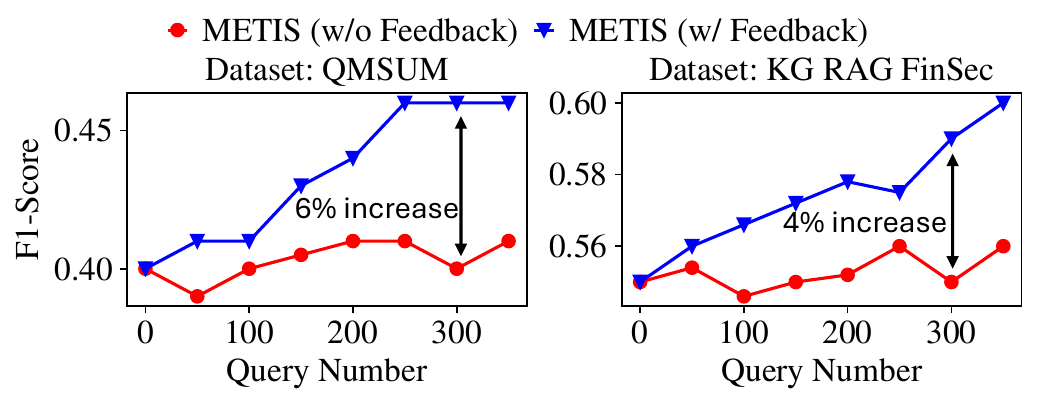}
     \vspace{-2pt}
    \caption{Improvement for \name using feedback from the output helps improve the F1-score by $4-6\%$.} 
    \label{fig:golden}
\end{figure}

\begin{figure}[t]
\centering
     \includegraphics[width=0.95\linewidth]{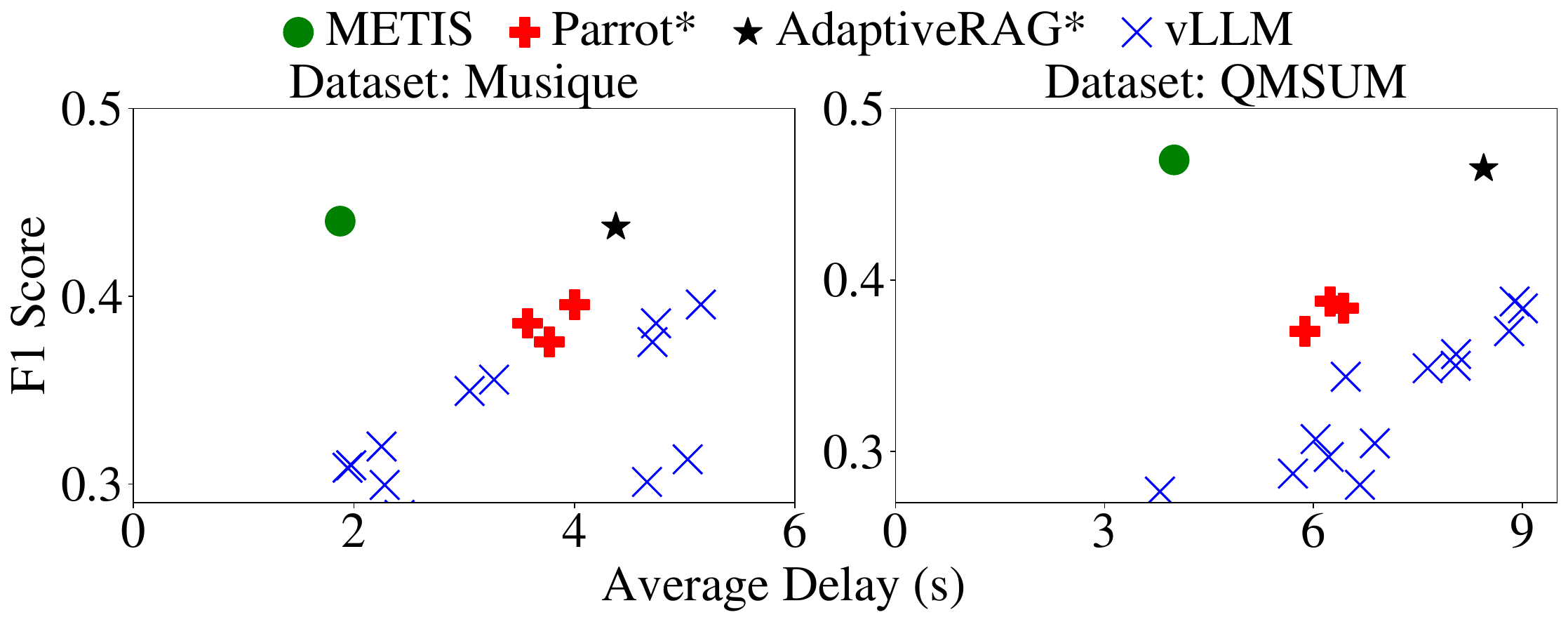}
     \vspace{-0.2cm}
    \caption{\name achieves lower delay by $2.1-2.4\times$ at the same quality even with a larger inference LLM.
    \vspace{-0.2cm}} 
    \label{fig:e2e_llama_infer}
\end{figure}

\begin{figure}[t]
    \centering
    \includegraphics[width=0.99\linewidth]{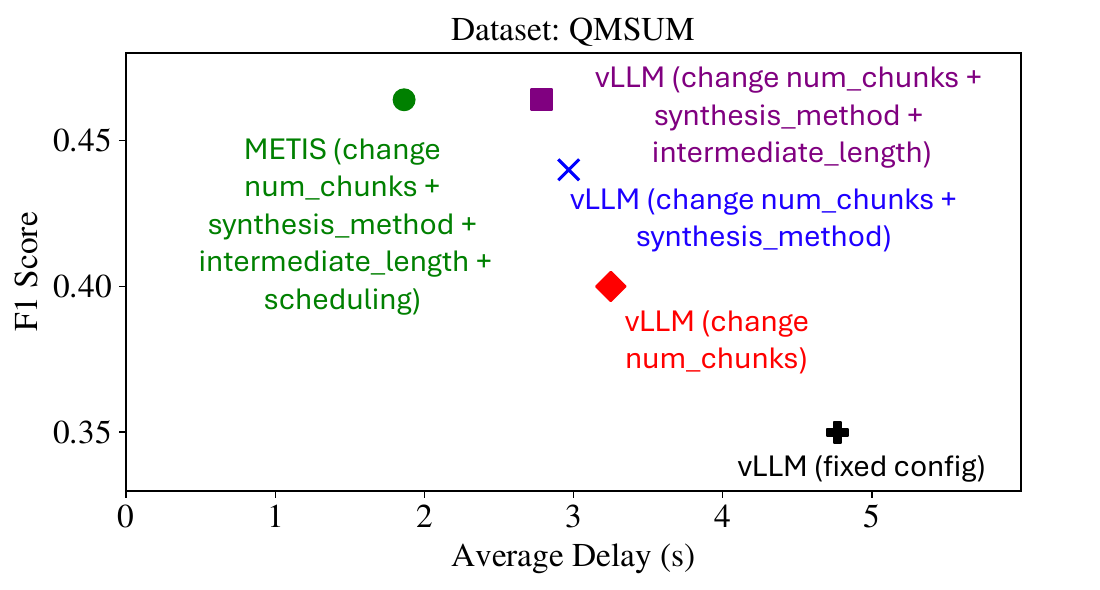}
    \caption{Breakdown analysis: By tuning more knobs in \name, we can see better quality-delay tradeoffs.\vspace{-0.2cm}}
    \label{fig:tuning}
\end{figure}

\begin{figure}[t]
\centering
     \includegraphics[width=0.9\linewidth]{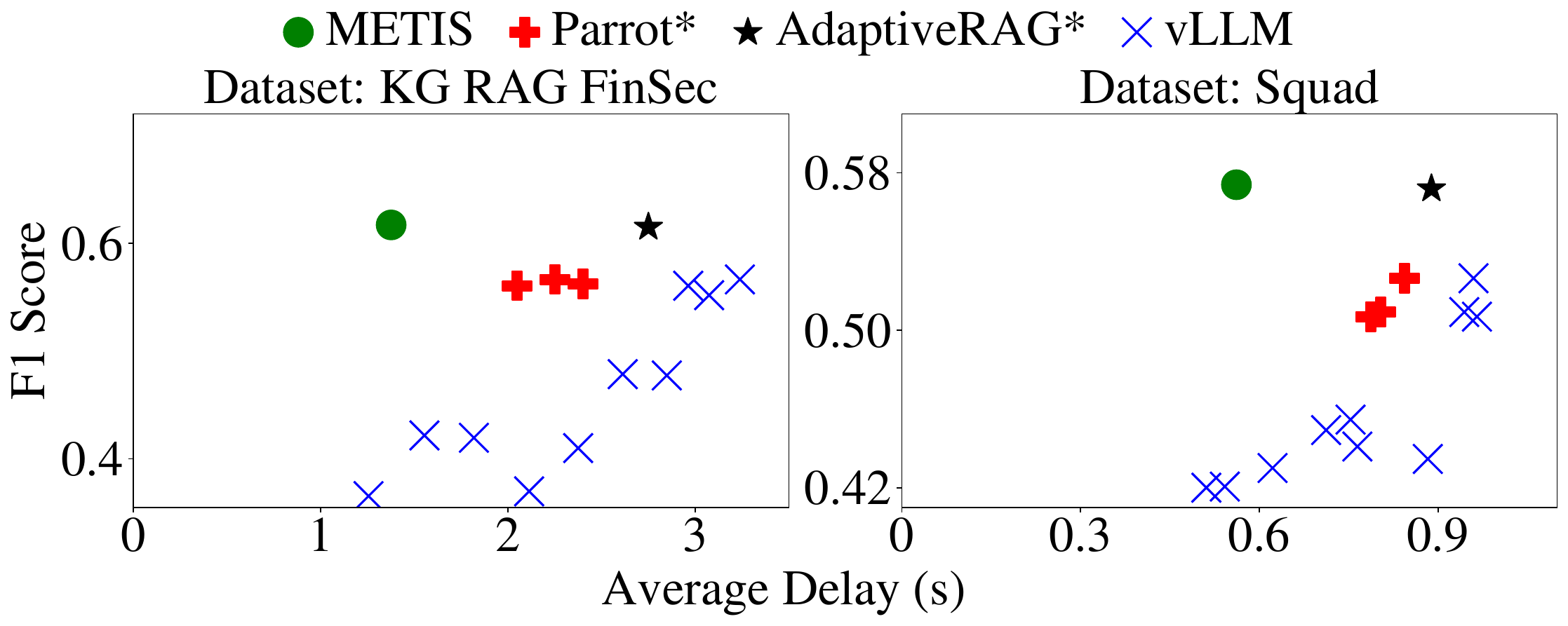}\vspace{-0.2cm}
    \caption{\name' performance gains remain substantial even with a smaller, open-source LLM profiler.\vspace{-0.2cm}}
    \label{fig:e2e_llama_lat}
\end{figure}

\begin{figure}[t]
    \centering
    \includegraphics[width=0.9\linewidth]{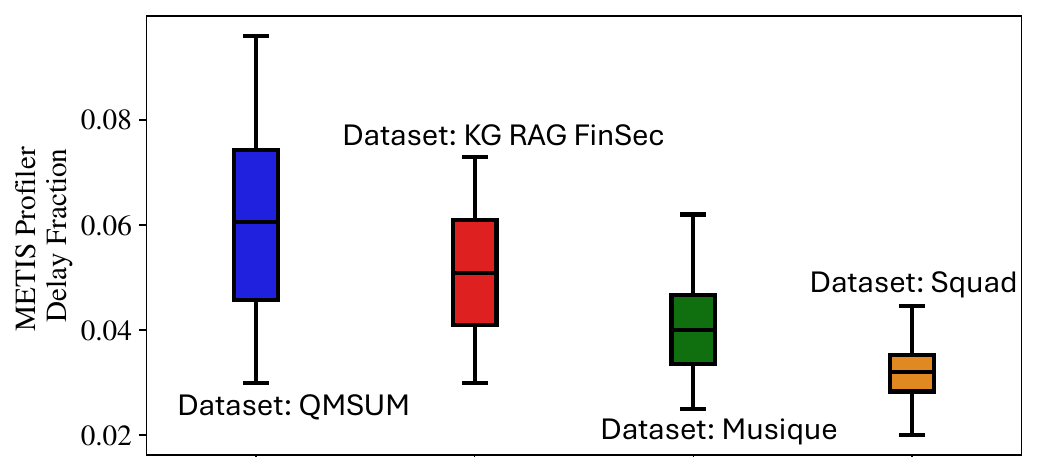}
    \caption{\name' profiler delay is at most 1/10th of end-to-end response delay across queries from all datasets.}
    \label{fig:profiler_fraction}
\end{figure}

\mypara{Cost saving} Figure \ref{fig:cost} shows \name (including its profiler) has significant lower dollar cost and higher F1-score, compared to choosing the best fixed configuration, with increasing model complexity. The cost of using a (LLama3-70B) inference model with vLLM and a fixed configuration is higher by $2.38\times$ times while also having a lower F1-score of $6.5\%$ times across datasets. Even more powerful inference models like GPT-4o fail to achieve the same F1-score with fixed configurations but have a much higher cost of $6.8\times$.

\mypara{Profiler feedback-based improvement}
In Figure \ref{fig:golden} we show the effect of the golden-configuration-based feedback to the profiler in order to improve its output. We use a 350 query sample for the QMSUM and KG RAG FinSec dataset as the workload. We see that with the feedback mechanism (blue line), the F1-score improves by $4-6 \%$ as compared to not having feedback (red line) from the outputs of the golden configuration. We ensure that the feedback mechanism cannot result in the output of very expensive configurations, as \name' joint scheduler will not pick increasingly expensive configurations based on the GPU resource constraint.

\tightsubsection{Sensitivity analysis}

\mypara{Changing the inference LLM} 
Figure \ref{fig:e2e_llama_infer} shows the outcome of changing the inference LLM to a larger LLM (Llama3.1-70B) on the Musique and QMSUM datasets. Even with a more powerful LLM, \name achieves $2.1-2.4\times$ lower delay than \emph{AdaptiveRAG*} at a similar F1-score. The best fixed-configuration baselines such as \emph{Parrot*} and \emph{vLLM} have a lower F1-score of $7-10\%$. In RAG, models mainly rely on the external context to answer the question instead of the model weights and we only get a $2\%$ improvement in F1-score compared to the smaller inference models.

\mypara{Incrementally tuning knobs in \name} In Figure \ref{fig:tuning}, we show the benefit we the improvement we get by incrementally adding more knobs to \name. We measure this for the QMSUM dataset with the original Mistral-7B-v3 model. We first only tune the \numchunks (red point). Progressively we tune the RAG-configuration knobs of  \synthesismeth and \interlen and scheduling. We achieve $5,4,3\%$ higher F1-Score compared to vLLM. Finally, by adding the scheduling, $2.8\times$ lower delay reduction in delay.



\mypara{Changing the profiler LLM} Figure \ref{fig:e2e_llama_lat} shows the effect of changing the LLM profiler from GPT-4o to a smaller Llama3.1-70B model. \name with the new profiler, still achieves  $1.4-2.1\times$ over \emph{AdaptiveRAG*} with a similar F1-score. Static configurations of \emph{Parrot*} and \emph{vLLM} which achieve similar delay, \name achieves $10-14\%$ higher F1-score.




\mypara{Delay overhead of \name's per-query profiling} We show the negligible delay overhead of using an LLM profiler within \name. Figure \ref{fig:profiler_fraction} shows the fraction of \name' profiler of the total end-to-end delay. Using the profiler at most adds $0.1$ fraction and in the average case only adds $0.03-0.06$ fraction to the total delay across queries from all datasets.

\begin{figure}[t]
    \centering
    \includegraphics[width=0.9\linewidth]{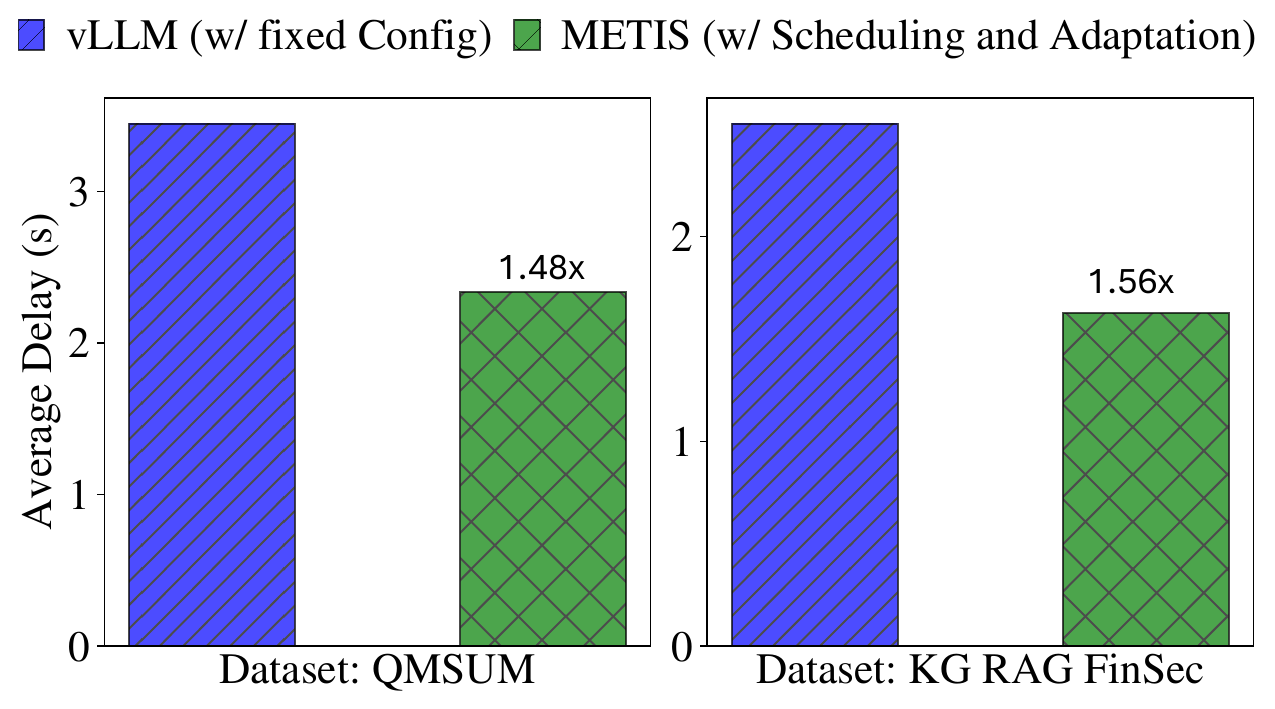}
    \tightcaption{\name' under low load without batching}
    \label{fig:delay_low_load}
\end{figure}

{\color{black}
\mypara{\name' performance under low load} In Figure \ref{fig:delay_low_load}, we evaluate \name under low load by sending queries sequentially, with every query sent after the previous query is completed. We compare with vLLM's fixed configuration, which achieves the highest quality (blue bar). \name uses its best-fit algorithm to pick the most expensive configuration from the pruned space of configurations. As \name only picks from configurations relevant to the query profile, it is still able to reduce delay by $1.48-1.56\times$ under low load.}

\tightsection{Related work}
\label{sec:related}


\mypara{Systems for serving RAG} Several systems have been proposed for RAG~\cite{lin2024parrotefficientservingllmbased,jin2024ragcacheefficientknowledgecaching,tan2024teolaendtoendoptimizationllmbased,martin2024llmproxyreducingcostaccess,kim2024effectschedulingpreemptionefficiency,jeong-etal-2024-adaptive,zhang2024pqcacheproductquantizationbasedkvcache,edge2024localglobalgraphrag,infercept,piperag,ralmspec} which focus on improving retrieval using complex, iterative retrieval algorithms or on serving model selection. \name can work in conjunction with such systems as \name focuses on optimizing quality and serving latency, independent of how the retrieval algorithm identifies chunks for retrieval.

\mypara{KV cache storage and retrieval} Storing and reusing KV cache across different requests have been commonly studied in recent work~\cite{10.1145/3651890.3672274,hu2024memservecontextcachingdisaggregated,liu2024minicachekvcachecompression,dong2024lesssynthesizingrecurrencekv,liu2024droidspeakenhancingcrossllmcommunication,pan2024marconi,infercept,cachedattention,pensieve,preble,zheng2024sglang,kwon2023efficient, jiang2025ragosystematicperformanceoptimization}. \name can work alongside these systems, where instead of retrieving chunks, it can retrieve the KV Caches for generating the output. In RAG, some additional optimizations are needed to combine KV Caches of different chunks that don't share a common prefix. This is important as the trivial concatenation of KV Caches loses important cross-attention and reasoning between chunks. These optimizations are enabled by KV Cache blending-based approaches~\cite{yao2024cacheblendfastlargelanguage,hu2024epicefficientpositionindependentcontext,cheng2024largelanguagemodelsneed,jin2024computeloadkvcache,wang2024modeltellsmergeadaptive,promptcache}. However RAG workloads have a large number of related contexts across queries and storing all the KV Cache is extremely expensive. We do not measure the KV Cache reuse ratio across queries  and leave it for future work.

\mypara{Prefill-Decode Optimizations} Several systems have proposed optimizations to speed-up prefill and decode for LLMs by leveraging unique properties of each phase~\cite{sarathi,distserve,10.1145/3694715.3695948,qin2024mooncakekvcachecentricdisaggregatedarchitecture,jiang2024neosavinggpumemory,apparate,powerinfer,nanoflow}. Notable techniques include \emph{chunked-prefill} which allows interleaving prefill and decode requests and \emph{disaggregated prefill} which separates compute nodes for prefill and decode. All of these optimizations enable faster generation speed but don't focus on generation quality. \name can be applied with such LLM serving systems optimizations.
\section{\textcolor{black}{Discussion and Limitations}}
\label{sec:discussion}

{\color{black}
While \name' profiler and configuration mapping algorithm is currently designed to work with commonly deployed RAG QA pipelines, it can be easily extended to generalize across new RAG configurations, workflows and domains.

\mypara{Agentic RAG} New research directions in RAG~\cite{zhang2024raftadaptinglanguagemodel, li2025agenticragdeepreasoning, guo2025reagannodeasagentreasoninggraphagentic} have developed pipelines with LLM agents, tool calling and deep \emph{chain-of-thought} to be used for RAG workloads. For an agentic workflow, a key extension for \name is to profile the query-complexity and break down a query into multiple sub-queries for planning (\eg how many sub-queries are needed becomes a new configuration knob). \name complements such workflows and can continue to perform the joint resource allocation for each sub-query.

\mypara{Multi-modal RAG} Emerging multi-modal LLMs has led to the need for multi-modal RAG~\cite{riedler2024textoptimizingragmultimodal, liu2025hmraghierarchicalmultiagentmultimodal, hu2025mragbenchvisioncentricevaluationretrievalaugmented} and \name can be extended to complement this. Using the natural-language profile of the query, \name can profile the different types of data to be retrieved. \eg A query might ask for a fact and a supporting image and this becomes a new configuration knob. Based on the properties of the new data retrieved, a configuration selection mapping rule can be added to decide which final RAG configuration should be chosen.

\mypara{Graph-based RAG} Another emerging area is GraphRAG~\cite{10.1145/3711896.3737012, edge2024localglobalgraphrag, zhou2025indepthanalysisgraphbasedrag} where retrieval is performed by choosing from hierarchical communities (\eg coarse grained data aggregation vs fine-grained facts). \name can complement such approaches by using the complexity profile of the query it generates, in order to choose the appropriate depth of community to use, and add this as a configuration knob. 


}

\section{Conclusion}

This paper introduces \name, the first system that focuses on optimizing the tradeoffs between response delay and generation quality in RAG, by 
by jointly scheduling RAG queries and adapting key configurations on a per-query basis. 
Evaluation on four datasets shows that \name outperforms the state-of-the-art, reducing generation latency by $1.64-2.54\times$ without compromising response quality. 

\section*{\textcolor{black}{Acknowledgment}}
\label{sec:acknowledgement}

\textcolor{black}{We thank all the anonymous reviewers and our shepherd Oana Balmau, for their insightful feedback and suggestions. The project is funded by NSF CNS-2146496, CNS-2131826, CNS-2313190, CNS-1901466, UChicago CERES Center and a Google Faculty Research Award.}

\bibliographystyle{plain}
\bibliography{main} 

\begin{thebibliography}{100}

\bibitem{rag_hyperparameter_tuning}
{Hyperparameter Optimization for RAG}.
\newblock
  \url{https://docs.llamaindex.ai/en/stable/examples/param_optimizer/param_optimizer/},
  2024.

\bibitem{infercept}
Reyna Abhyankar, Zijian He, Vikranth Srivatsa, Hao Zhang, and Yiying Zhang.
\newblock Infercept: Efficient intercept support for augmented large language
  model inference.
\newblock In {\em Forty-first International Conference on Machine Learning}.

\bibitem{sarathi}
Amey Agrawal, Nitin Kedia, Ashish Panwar, Jayashree Mohan, Nipun Kwatra,
  Bhargav Gulavani, Alexey Tumanov, and Ramachandran Ramjee.
\newblock Taming {Throughput-Latency} tradeoff in {LLM} inference with
  {Sarathi-Serve}.
\newblock In {\em 18th USENIX Symposium on Operating Systems Design and
  Implementation (OSDI 24)}, pages 117--134, Santa Clara, CA, July 2024. USENIX
  Association.

\bibitem{cohere_embed}
Cohere AI.
\newblock Cohere: Cutting-edge gen ai, 2023.

\bibitem{Ansel_PyTorch_2_Faster_2024}
Jason et~al. Ansel.
\newblock {PyTorch 2: Faster Machine Learning Through Dynamic Python Bytecode
  Transformation and Graph Compilation}.
\newblock In {\em 29th ACM International Conference on Architectural Support
  for Programming Languages and Operating Systems, Volume 2 (ASPLOS '24)}. ACM,
  April 2024.

\bibitem{acl_rag_tutorial}
Akari Asai, Sewon Min, Zexuan Zhong, and Danqi Chen.
\newblock Retrieval-based language models and applications.
\newblock In {\em Proceedings of the 61st Annual Meeting of the Association for
  Computational Linguistics (Volume 6: Tutorial Abstracts)}, pages 41--46,
  2023.

\bibitem{balaguer2024ragvsfinetuningpipelines}
Angels Balaguer, Vinamra Benara, Renato~Luiz de~Freitas~Cunha, Roberto
  de~M.~Estevão~Filho, Todd Hendry, Daniel Holstein, Jennifer Marsman, Nick
  Mecklenburg, Sara Malvar, Leonardo~O. Nunes, Rafael Padilha, Morris Sharp,
  Bruno Silva, Swati Sharma, Vijay Aski, and Ranveer Chandra.
\newblock Rag vs fine-tuning: Pipelines, tradeoffs, and a case study on
  agriculture, 2024.

\bibitem{Chase_LangChain_2022}
Harrison Chase.
\newblock {LangChain}, October 2022.

\bibitem{cheng2024largelanguagemodelsneed}
Yihua Cheng, Kuntai Du, Jiayi Yao, and Junchen Jiang.
\newblock Do large language models need a content delivery network?, 2024.

\bibitem{10.1145/3626772.3657834}
Florin Cuconasu, Giovanni Trappolini, Federico Siciliano, Simone Filice, Cesare
  Campagnano, Yoelle Maarek, Nicola Tonellotto, and Fabrizio Silvestri.
\newblock The power of noise: Redefining retrieval for rag systems.
\newblock In {\em Proceedings of the 47th International ACM SIGIR Conference on
  Research and Development in Information Retrieval}, SIGIR '24, page
  719–729, New York, NY, USA, 2024. Association for Computing Machinery.

\bibitem{apparate}
Yinwei Dai, Rui Pan, Anand Iyer, Kai Li, and Ravi Netravali.
\newblock Apparate: Rethinking early exits to tame latency-throughput tensions
  in ml serving.
\newblock In {\em Proceedings of the ACM SIGOPS 30th Symposium on Operating
  Systems Principles}, pages 607--623, 2024.

\bibitem{devlin2019bertpretrainingdeepbidirectional}
Jacob Devlin, Ming-Wei Chang, Kenton Lee, and Kristina Toutanova.
\newblock Bert: Pre-training of deep bidirectional transformers for language
  understanding, 2019.

\bibitem{ding2024hybridllmcostefficientqualityaware}
Dujian Ding, Ankur Mallick, Chi Wang, Robert Sim, Subhabrata Mukherjee, Victor
  Ruhle, Laks V.~S. Lakshmanan, and Ahmed~Hassan Awadallah.
\newblock Hybrid llm: Cost-efficient and quality-aware query routing, 2024.

\bibitem{dong2024lesssynthesizingrecurrencekv}
Harry Dong, Xinyu Yang, Zhenyu Zhang, Zhangyang Wang, Yuejie Chi, and Beidi
  Chen.
\newblock Get more with less: Synthesizing recurrence with kv cache compression
  for efficient llm inference, 2024.

\bibitem{10.1145/3637528.3671445}
Jos\'{e}~Cassio dos Santos~Junior, Rachel Hu, Richard Song, and Yunfei Bai.
\newblock Domain-driven llm development: Insights into rag and fine-tuning
  practices.
\newblock In {\em Proceedings of the 30th ACM SIGKDD Conference on Knowledge
  Discovery and Data Mining}, KDD '24, page 6416–6417, New York, NY, USA,
  2024. Association for Computing Machinery.

\bibitem{douze2024faiss}
Matthijs Douze, Alexandr Guzhva, Chengqi Deng, Jeff Johnson, Gergely Szilvasy,
  Pierre-Emmanuel Mazaré, Maria Lomeli, Lucas Hosseini, and Hervé Jégou.
\newblock The faiss library.
\newblock 2024.

\bibitem{edge2024localglobalgraphrag}
Darren Edge, Ha~Trinh, Newman Cheng, Joshua Bradley, Alex Chao, Apurva Mody,
  Steven Truitt, and Jonathan Larson.
\newblock From local to global: A graph rag approach to query-focused
  summarization, 2024.

\bibitem{neelakantan2022textcodeembeddingscontrastive}
Arvind~Neelakantan et~al.
\newblock Text and code embeddings by contrastive pre-training, 2022.

\bibitem{mteb_benchmark}
Hugging Face.
\newblock Mteb: Massive text embedding benchmark, 2024.

\bibitem{10.1145/3637528.3671470}
Wenqi Fan, Yujuan Ding, Liangbo Ning, Shijie Wang, Hengyun Li, Dawei Yin,
  Tat-Seng Chua, and Qing Li.
\newblock A survey on rag meeting llms: Towards retrieval-augmented large
  language models.
\newblock In {\em Proceedings of the 30th ACM SIGKDD Conference on Knowledge
  Discovery and Data Mining}, KDD '24, page 6491–6501, New York, NY, USA,
  2024. Association for Computing Machinery.

\bibitem{autorag_hp}
Jia Fu, Xiaoting Qin, Fangkai Yang, Lu~Wang, Jue Zhang, Qingwei Lin, Yubo Chen,
  Dongmei Zhang, Saravan Rajmohan, and Qi~Zhang.
\newblock Autorag-hp: Automatic online hyper-parameter tuning for
  retrieval-augmented generation.
\newblock {\em arXiv preprint arXiv:2406.19251}, 2024.

\bibitem{cachedattention}
Bin Gao, Zhuomin He, Puru Sharma, Qingxuan Kang, Djordje Jevdjic, Junbo Deng,
  Xingkun Yang, Zhou Yu, and Pengfei Zuo.
\newblock Cost-efficient large language model serving for multi-turn
  conversations with cachedattention.
\newblock In {\em 2024 USENIX Annual Technical Conference (USENIX ATC 24)},
  pages 111--126, 2024.

\bibitem{Gao_2022}
Luyu Gao and Jamie Callan.
\newblock Long document re-ranking with modular re-ranker.
\newblock In {\em Proceedings of the 45th International ACM SIGIR Conference on
  Research and Development in Information Retrieval}, SIGIR ’22, page
  2371–2376. ACM, July 2022.

\bibitem{ragblog1}
Aude Genevay.
\newblock From rag to fabric: Lessons learned from building real-world rags at
  genaiic – part 1.
\newblock Technical report, AWS, 2024.

\bibitem{geng-etal-2024-survey}
Jiahui Geng, Fengyu Cai, Yuxia Wang, Heinz Koeppl, Preslav Nakov, and Iryna
  Gurevych.
\newblock A survey of confidence estimation and calibration in large language
  models.
\newblock In Kevin Duh, Helena Gomez, and Steven Bethard, editors, {\em
  Proceedings of the 2024 Conference of the North American Chapter of the
  Association for Computational Linguistics: Human Language Technologies
  (Volume 1: Long Papers)}, pages 6577--6595, Mexico City, Mexico, June 2024.
  Association for Computational Linguistics.

\bibitem{promptcache}
In~Gim, Guojun Chen, Seung-seob Lee, Nikhil Sarda, Anurag Khandelwal, and Lin
  Zhong.
\newblock Prompt cache: Modular attention reuse for low-latency inference.
\newblock {\em Proceedings of Machine Learning and Systems}, 6:325--338, 2024.

\bibitem{guo2025reagannodeasagentreasoninggraphagentic}
Minghao Guo, Xi~Zhu, Jingyuan Huang, Kai Mei, and Yongfeng Zhang.
\newblock Reagan: Node-as-agent-reasoning graph agentic network, 2025.

\bibitem{guo2024lightragsimplefastretrievalaugmented}
Zirui Guo, Lianghao Xia, Yanhua Yu, Tu~Ao, and Chao Huang.
\newblock Lightrag: Simple and fast retrieval-augmented generation, 2024.

\bibitem{hsieh2024rulerwhatsrealcontext}
Cheng-Ping Hsieh, Simeng Sun, Samuel Kriman, Shantanu Acharya, Dima Rekesh, Fei
  Jia, Yang Zhang, and Boris Ginsburg.
\newblock Ruler: What's the real context size of your long-context language
  models?, 2024.

\bibitem{hu2024memservecontextcachingdisaggregated}
Cunchen Hu, Heyang Huang, Junhao Hu, Jiang Xu, Xusheng Chen, Tao Xie, Chenxi
  Wang, Sa~Wang, Yungang Bao, Ninghui Sun, and Yizhou Shan.
\newblock Memserve: Context caching for disaggregated llm serving with elastic
  memory pool, 2024.

\bibitem{hu2024epicefficientpositionindependentcontext}
Junhao Hu, Wenrui Huang, Haoyi Wang, Weidong Wang, Tiancheng Hu, Qin Zhang, Hao
  Feng, Xusheng Chen, Yizhou Shan, and Tao Xie.
\newblock Epic: Efficient position-independent context caching for serving
  large language models, 2024.

\bibitem{hu2024routerbenchbenchmarkmultillmrouting}
Qitian~Jason Hu, Jacob Bieker, Xiuyu Li, Nan Jiang, Benjamin Keigwin, Gaurav
  Ranganath, Kurt Keutzer, and Shriyash~Kaustubh Upadhyay.
\newblock Routerbench: A benchmark for multi-llm routing system, 2024.

\bibitem{hu2025mragbenchvisioncentricevaluationretrievalaugmented}
Wenbo Hu, Jia-Chen Gu, Zi-Yi Dou, Mohsen Fayyaz, Pan Lu, Kai-Wei Chang, and
  Nanyun Peng.
\newblock Mrag-bench: Vision-centric evaluation for retrieval-augmented
  multimodal models, 2025.

\bibitem{10.1145/3711896.3737012}
Yiqian Huang, Shiqi Zhang, and Xiaokui Xiao.
\newblock Ket-rag: A cost-efficient multi-granular indexing framework for
  graph-rag.
\newblock In {\em Proceedings of the 31st ACM SIGKDD Conference on Knowledge
  Discovery and Data Mining V.2}, KDD '25, page 1003–1012, New York, NY, USA,
  2025. Association for Computing Machinery.

\bibitem{jeong-etal-2024-adaptive}
Soyeong Jeong, Jinheon Baek, Sukmin Cho, Sung~Ju Hwang, and Jong Park.
\newblock Adaptive-{RAG}: Learning to adapt retrieval-augmented large language
  models through question complexity.
\newblock In Kevin Duh, Helena Gomez, and Steven Bethard, editors, {\em
  Proceedings of the 2024 Conference of the North American Chapter of the
  Association for Computational Linguistics: Human Language Technologies
  (Volume 1: Long Papers)}, pages 7036--7050, Mexico City, Mexico, June 2024.
  Association for Computational Linguistics.

\bibitem{jiang2025ragosystematicperformanceoptimization}
Wenqi Jiang, Suvinay Subramanian, Cat Graves, Gustavo Alonso, Amir
  Yazdanbakhsh, and Vidushi Dadu.
\newblock Rago: Systematic performance optimization for retrieval-augmented
  generation serving, 2025.

\bibitem{piperag}
Wenqi Jiang, Shuai Zhang, Boran Han, Jie Wang, Bernie Wang, and Tim Kraska.
\newblock Piperag: Fast retrieval-augmented generation via algorithm-system
  co-design.
\newblock {\em arXiv preprint arXiv:2403.05676}, 2024.

\bibitem{jiang2024neosavinggpumemory}
Xuanlin Jiang, Yang Zhou, Shiyi Cao, Ion Stoica, and Minlan Yu.
\newblock Neo: Saving gpu memory crisis with cpu offloading for online llm
  inference, 2024.

\bibitem{jiang2023activeretrievalaugmentedgeneration}
Zhengbao Jiang, Frank~F. Xu, Luyu Gao, Zhiqing Sun, Qian Liu, Jane Dwivedi-Yu,
  Yiming Yang, Jamie Callan, and Graham Neubig.
\newblock Active retrieval augmented generation, 2023.

\bibitem{jin2024ragcacheefficientknowledgecaching}
Chao Jin, Zili Zhang, Xuanlin Jiang, Fangyue Liu, Xin Liu, Xuanzhe Liu, and Xin
  Jin.
\newblock Ragcache: Efficient knowledge caching for retrieval-augmented
  generation, 2024.

\bibitem{jin2024computeloadkvcache}
Shuowei Jin, Xueshen Liu, Qingzhao Zhang, and Z.~Morley Mao.
\newblock Compute or load kv cache? why not both?, 2024.

\bibitem{kim2024autoragautomatedframeworkoptimization}
Dongkyu Kim, Byoungwook Kim, Donggeon Han, and Matouš Eibich.
\newblock Autorag: Automated framework for optimization of retrieval augmented
  generation pipeline, 2024.

\bibitem{kim2024effectschedulingpreemptionefficiency}
Kyoungmin Kim, Kijae Hong, Caglar Gulcehre, and Anastasia Ailamaki.
\newblock The effect of scheduling and preemption on the efficiency of llm
  inference serving, 2024.

\bibitem{kwon2023efficient}
Woosuk Kwon, Zhuohan Li, Siyuan Zhuang, Ying Sheng, Lianmin Zheng, Cody~Hao Yu,
  Joseph~E. Gonzalez, Hao Zhang, and Ion Stoica.
\newblock Efficient memory management for large language model serving with
  pagedattention.
\newblock In {\em Proceedings of the ACM SIGOPS 29th Symposium on Operating
  Systems Principles}, 2023.

\bibitem{leng2024longcontextragperformance}
Quinn Leng, Jacob Portes, Sam Havens, Matei Zaharia, and Michael Carbin.
\newblock Long context rag performance of large language models, 2024.

\bibitem{li2025agenticragdeepreasoning}
Yangning Li, Weizhi Zhang, Yuyao Yang, Wei-Chieh Huang, Yaozu Wu, Junyu Luo,
  Yuanchen Bei, Henry~Peng Zou, Xiao Luo, Yusheng Zhao, Chunkit Chan, Yankai
  Chen, Zhongfen Deng, Yinghui Li, Hai-Tao Zheng, Dongyuan Li, Renhe Jiang,
  Ming Zhang, Yangqiu Song, and Philip~S. Yu.
\newblock Towards agentic rag with deep reasoning: A survey of rag-reasoning
  systems in llms, 2025.

\bibitem{li2024retrievalaugmentedgenerationlongcontext}
Zhuowan Li, Cheng Li, Mingyang Zhang, Qiaozhu Mei, and Michael Bendersky.
\newblock Retrieval augmented generation or long-context llms? a comprehensive
  study and hybrid approach, 2024.

\bibitem{lin2024parrotefficientservingllmbased}
Chaofan Lin, Zhenhua Han, Chengruidong Zhang, Yuqing Yang, Fan Yang, Chen Chen,
  and Lili Qiu.
\newblock Parrot: Efficient serving of llm-based applications with semantic
  variable, 2024.

\bibitem{lin2025teleragefficientretrievalaugmentedgeneration}
Chien-Yu Lin, Keisuke Kamahori, Yiyu Liu, Xiaoxiang Shi, Madhav Kashyap, Yile
  Gu, Rulin Shao, Zihao Ye, Kan Zhu, Stephanie Wang, Arvind Krishnamurthy,
  Rohan Kadekodi, Luis Ceze, and Baris Kasikci.
\newblock Telerag: Efficient retrieval-augmented generation inference with
  lookahead retrieval, 2025.

\bibitem{liu2024minicachekvcachecompression}
Akide Liu, Jing Liu, Zizheng Pan, Yefei He, Gholamreza Haffari, and Bohan
  Zhuang.
\newblock Minicache: Kv cache compression in depth dimension for large language
  models, 2024.

\bibitem{lost_in_the_middle}
Nelson~F Liu, Kevin Lin, John Hewitt, Ashwin Paranjape, Michele Bevilacqua,
  Fabio Petroni, and Percy Liang.
\newblock Lost in the middle: How language models use long contexts.
\newblock {\em Transactions of the Association for Computational Linguistics},
  12:157--173, 2024.

\bibitem{liu2025hmraghierarchicalmultiagentmultimodal}
Pei Liu, Xin Liu, Ruoyu Yao, Junming Liu, Siyuan Meng, Ding Wang, and Jun Ma.
\newblock Hm-rag: Hierarchical multi-agent multimodal retrieval augmented
  generation, 2025.

\bibitem{liu2024droidspeakenhancingcrossllmcommunication}
Yuhan Liu, Esha Choukse, Shan Lu, Junchen Jiang, and Madan Musuvathi.
\newblock Droidspeak: Enhancing cross-llm communication, 2024.

\bibitem{10.1145/3651890.3672274}
Yuhan Liu, Hanchen Li, Yihua Cheng, Siddhant Ray, Yuyang Huang, Qizheng Zhang,
  Kuntai Du, Jiayi Yao, Shan Lu, Ganesh Ananthanarayanan, Michael Maire, Henry
  Hoffmann, Ari Holtzman, and Junchen Jiang.
\newblock Cachegen: Kv cache compression and streaming for fast large language
  model serving.
\newblock In {\em Proceedings of the ACM SIGCOMM 2024 Conference}, ACM SIGCOMM
  '24, page 38–56, New York, NY, USA, 2024. Association for Computing
  Machinery.

\bibitem{kgrag}
LlamaHub.
\newblock Docugami knowledge graph retrieval augmented generation (kg-rag)
  datasets, 2024.

\bibitem{ma2023queryrewritingretrievalaugmentedlarge}
Xinbei Ma, Yeyun Gong, Pengcheng He, Hai Zhao, and Nan Duan.
\newblock Query rewriting for retrieval-augmented large language models, 2023.

\bibitem{ma2023zeroshotlistwisedocumentreranking}
Xueguang Ma, Xinyu Zhang, Ronak Pradeep, and Jimmy Lin.
\newblock Zero-shot listwise document reranking with a large language model,
  2023.

\bibitem{mao-etal-2024-rag}
Kelong Mao, Zheng Liu, Hongjin Qian, Fengran Mo, Chenlong Deng, and Zhicheng
  Dou.
\newblock {RAG}-studio: Towards in-domain adaptation of retrieval augmented
  generation through self-alignment.
\newblock In Yaser Al-Onaizan, Mohit Bansal, and Yun-Nung Chen, editors, {\em
  Findings of the Association for Computational Linguistics: EMNLP 2024}, pages
  725--735, Miami, Florida, USA, November 2024. Association for Computational
  Linguistics.

\bibitem{martin2024llmproxyreducingcostaccess}
Noah Martin, Abdullah~Bin Faisal, Hiba Eltigani, Rukhshan Haroon, Swaminathan
  Lamelas, and Fahad Dogar.
\newblock Llmproxy: Reducing cost to access large language models, 2024.

\bibitem{ragblog2}
Rick Merritt.
\newblock What is retrieval-augmented generation, aka rag?
\newblock Technical report, NVIDIA, 2024.

\bibitem{mohammadshahi2024routoolearningroutelarge}
Alireza Mohammadshahi, Arshad~Rafiq Shaikh, and Majid Yazdani.
\newblock Routoo: Learning to route to large language models effectively, 2024.

\bibitem{mombaerts2024metaknowledgeretrievalaugmented}
Laurent Mombaerts, Terry Ding, Adi Banerjee, Florian Felice, Jonathan Taws, and
  Tarik Borogovac.
\newblock Meta knowledge for retrieval augmented large language models, 2024.

\bibitem{nguyen2024enhancingqadomain}
Zooey Nguyen, Anthony Annunziata, Vinh Luong, Sang Dinh, Quynh Le, Anh~Hai Ha,
  Chanh Le, Hong~An Phan, Shruti Raghavan, and Christopher Nguyen.
\newblock Enhancing q\&a with domain-specific fine-tuning and iterative
  reasoning: A comparative study, 2024.

\bibitem{ong2024routellmlearningroutellms}
Isaac Ong, Amjad Almahairi, Vincent Wu, Wei-Lin Chiang, Tianhao Wu, Joseph~E.
  Gonzalez, M~Waleed Kadous, and Ion Stoica.
\newblock Routellm: Learning to route llms with preference data, 2024.

\bibitem{ong2025routellmlearningroutellms}
Isaac Ong, Amjad Almahairi, Vincent Wu, Wei-Lin Chiang, Tianhao Wu, Joseph~E.
  Gonzalez, M~Waleed Kadous, and Ion Stoica.
\newblock Routellm: Learning to route llms with preference data, 2025.

\bibitem{OpenAIAPI}
OpenAI.
\newblock Openai api, 2023.

\bibitem{ouyang2024contextaware}
Yang Ouyang, Tong Yu, and Wenchu Wang.
\newblock Context-aware chatbot extension leveraging {HTML} data and
  retrieval-augmented generation ({RAG}).
\newblock In {\em Submitted to Tsinghua University Course: Advanced Machine
  Learning}, 2024.
\newblock under review.

\bibitem{pan2024marconi}
Rui Pan, Zhuang Wang, Zhen Jia, Can Karakus, Luca Zancato, Tri Dao, Yida Wang,
  and Ravi Netravali.
\newblock Marconi: Prefix caching for the era of hybrid llms.
\newblock {\em arXiv preprint arXiv:2411.19379}, 2024.

\bibitem{qian2024memoragmovingnextgenrag}
Hongjin Qian, Peitian Zhang, Zheng Liu, Kelong Mao, and Zhicheng Dou.
\newblock Memorag: Moving towards next-gen rag via memory-inspired knowledge
  discovery, 2024.

\bibitem{qin2024mooncakekvcachecentricdisaggregatedarchitecture}
Ruoyu Qin, Zheming Li, Weiran He, Mingxing Zhang, Yongwei Wu, Weimin Zheng, and
  Xinran Xu.
\newblock Mooncake: A kvcache-centric disaggregated architecture for llm
  serving, 2024.

\bibitem{squad}
Pranav Rajpurkar, Jian Zhang, Konstantin Lopyrev, and Percy Liang.
\newblock {SQ}u{AD}: 100,000+ questions for machine comprehension of text.
\newblock In Jian Su, Kevin Duh, and Xavier Carreras, editors, {\em Proceedings
  of the 2016 Conference on Empirical Methods in Natural Language Processing},
  pages 2383--2392, Austin, Texas, November 2016. Association for Computational
  Linguistics.

\bibitem{reimers2019sentencebertsentenceembeddingsusing}
Nils Reimers and Iryna Gurevych.
\newblock Sentence-bert: Sentence embeddings using siamese bert-networks, 2019.

\bibitem{rag_industry}
Grand~View Research.
\newblock Retrieval augmented generation market size, share and trend analysis
  report by function (document retrieval, recommendation engines), by
  application (content generation), by deployment (cloud, on-premises), by end
  use, by region, and segment forecasts, 2025 - 2030, 2024.

\bibitem{riedler2024textoptimizingragmultimodal}
Monica Riedler and Stefan Langer.
\newblock Beyond text: Optimizing rag with multimodal inputs for industrial
  applications, 2024.

\bibitem{ru2024ragcheckerfinegrainedframeworkdiagnosing}
Dongyu Ru, Lin Qiu, Xiangkun Hu, Tianhang Zhang, Peng Shi, Shuaichen Chang,
  Cheng Jiayang, Cunxiang Wang, Shichao Sun, Huanyu Li, Zizhao Zhang, Binjie
  Wang, Jiarong Jiang, Tong He, Zhiguo Wang, Pengfei Liu, Yue Zhang, and Zheng
  Zhang.
\newblock Ragchecker: A fine-grained framework for diagnosing
  retrieval-augmented generation, 2024.

\bibitem{shahout2024fastinferenceaugmentedlarge}
Rana Shahout, Cong Liang, Shiji Xin, Qianru Lao, Yong Cui, Minlan Yu, and
  Michael Mitzenmacher.
\newblock Fast inference for augmented large language models, 2024.

\bibitem{NEURIPS2024_a5d8aba2}
Rulin Shao, Jacqueline He, Akari Asai, Weijia Shi, Tim Dettmers, Sewon Min,
  Luke Zettlemoyer, and Pang~Wei Koh.
\newblock Scaling retrieval-based language models with a trillion-token
  datastore.
\newblock In A.~Globerson, L.~Mackey, D.~Belgrave, A.~Fan, U.~Paquet,
  J.~Tomczak, and C.~Zhang, editors, {\em Advances in Neural Information
  Processing Systems}, volume~37, pages 91260--91299. Curran Associates, Inc.,
  2024.

\bibitem{simon2024methodologyevaluatingragsystems}
Sebastian Simon, Alina Mailach, Johannes Dorn, and Norbert Siegmund.
\newblock A methodology for evaluating rag systems: A case study on
  configuration dependency validation, 2024.

\bibitem{singh2025agenticretrievalaugmentedgenerationsurvey}
Aditi Singh, Abul Ehtesham, Saket Kumar, and Tala~Talaei Khoei.
\newblock Agentic retrieval-augmented generation: A survey on agentic rag,
  2025.

\bibitem{powerinfer}
Yixin Song, Zeyu Mi, Haotong Xie, and Haibo Chen.
\newblock Powerinfer: Fast large language model serving with a consumer-grade
  gpu.
\newblock In {\em Proceedings of the ACM SIGOPS 30th Symposium on Operating
  Systems Principles}, pages 590--606, 2024.

\bibitem{preble}
Vikranth Srivatsa, Zijian He, Reyna Abhyankar, Dongming Li, and Yiying Zhang.
\newblock Preble: Efficient distributed prompt scheduling for llm serving.
\newblock 2024.

\bibitem{tan2024teolaendtoendoptimizationllmbased}
Xin Tan, Yimin Jiang, Yitao Yang, and Hong Xu.
\newblock Teola: Towards end-to-end optimization of llm-based applications,
  2024.

\bibitem{tang2024mbaragbanditapproachadaptive}
Xiaqiang Tang, Qiang Gao, Jian Li, Nan Du, Qi~Li, and Sihong Xie.
\newblock Mba-rag: a bandit approach for adaptive retrieval-augmented
  generation through question complexity, 2024.

\bibitem{musique}
Harsh Trivedi, Niranjan Balasubramanian, Tushar Khot, and Ashish Sabharwal.
\newblock {M}u{S}i{Q}ue: Multihop questions via single-hop question
  composition.
\newblock {\em Transactions of the Association for Computational Linguistics},
  10:539--554, 2022.

\bibitem{wang-etal-2024-searching}
Xiaohua Wang, Zhenghua Wang, Xuan Gao, Feiran Zhang, Yixin Wu, Zhibo Xu,
  Tianyuan Shi, Zhengyuan Wang, Shizheng Li, Qi~Qian, Ruicheng Yin, Changze Lv,
  Xiaoqing Zheng, and Xuanjing Huang.
\newblock Searching for best practices in retrieval-augmented generation.
\newblock In Yaser Al-Onaizan, Mohit Bansal, and Yun-Nung Chen, editors, {\em
  Proceedings of the 2024 Conference on Empirical Methods in Natural Language
  Processing}, pages 17716--17736, Miami, Florida, USA, November 2024.
  Association for Computational Linguistics.

\bibitem{wang2024modeltellsmergeadaptive}
Zheng Wang, Boxiao Jin, Zhongzhi Yu, and Minjia Zhang.
\newblock Model tells you where to merge: Adaptive kv cache merging for llms on
  long-context tasks, 2024.

\bibitem{Wolf_Transformers_State-of-the-Art_Natural_2020}
Thomas Wolf, Lysandre Debut, Victor Sanh, Julien Chaumond, Clement Delangue,
  Anthony Moi, Perric Cistac, Clara Ma, Yacine Jernite, Julien Plu, Canwen Xu,
  Teven Le~Scao, Sylvain Gugger, Mariama Drame, Quentin Lhoest, and
  Alexander~M. Rush.
\newblock {Transformers: State-of-the-Art Natural Language Processing}.
\newblock pages 38--45. Association for Computational Linguistics, October
  2020.

\bibitem{10.1145/3694715.3695948}
Bingyang Wu, Shengyu Liu, Yinmin Zhong, Peng Sun, Xuanzhe Liu, and Xin Jin.
\newblock Loongserve: Efficiently serving long-context large language models
  with elastic sequence parallelism.
\newblock In {\em Proceedings of the ACM SIGOPS 30th Symposium on Operating
  Systems Principles}, SOSP '24, page 640–654, New York, NY, USA, 2024.
  Association for Computing Machinery.

\bibitem{xie2024weknowragadaptiveapproachretrievalaugmented}
Weijian Xie, Xuefeng Liang, Yuhui Liu, Kaihua Ni, Hong Cheng, and Zetian Hu.
\newblock Weknow-rag: An adaptive approach for retrieval-augmented generation
  integrating web search and knowledge graphs, 2024.

\bibitem{xiong2024can}
Miao Xiong, Zhiyuan Hu, Xinyang Lu, YIFEI LI, Jie Fu, Junxian He, and Bryan
  Hooi.
\newblock Can {LLM}s express their uncertainty? an empirical evaluation of
  confidence elicitation in {LLM}s.
\newblock In {\em The Twelfth International Conference on Learning
  Representations}, 2024.

\bibitem{yao2024cacheblendfastlargelanguage}
Jiayi Yao, Hanchen Li, Yuhan Liu, Siddhant Ray, Yihua Cheng, Qizheng Zhang,
  Kuntai Du, Shan Lu, and Junchen Jiang.
\newblock Cacheblend: Fast large language model serving for rag with cached
  knowledge fusion, 2024.

\bibitem{pensieve}
Lingfan Yu and Jinyang Li.
\newblock Stateful large language model serving with pensieve.
\newblock {\em arXiv preprint arXiv:2312.05516}, 2023.

\bibitem{zhang2024pqcacheproductquantizationbasedkvcache}
Hailin Zhang, Xiaodong Ji, Yilin Chen, Fangcheng Fu, Xupeng Miao, Xiaonan Nie,
  Weipeng Chen, and Bin Cui.
\newblock Pqcache: Product quantization-based kvcache for long context llm
  inference, 2024.

\bibitem{298701}
Qizheng Zhang, Ali Imran, Enkeleda Bardhi, Tushar Swamy, Nathan Zhang, Muhammad
  Shahbaz, and Kunle Olukotun.
\newblock Caravan: Practical online learning of {In-Network} {ML} models with
  labeling agents.
\newblock In {\em 18th USENIX Symposium on Operating Systems Design and
  Implementation (OSDI 24)}, pages 325--345, Santa Clara, CA, July 2024. USENIX
  Association.

\bibitem{zhang2024raftadaptinglanguagemodel}
Tianjun Zhang, Shishir~G. Patil, Naman Jain, Sheng Shen, Matei Zaharia, Ion
  Stoica, and Joseph~E. Gonzalez.
\newblock Raft: Adapting language model to domain specific rag, 2024.

\bibitem{ralmspec}
Zhihao Zhang, Alan Zhu, Lijie Yang, Yihua Xu, Lanting Li, Phitchaya~Mangpo
  Phothilimthana, and Zhihao Jia.
\newblock Accelerating iterative retrieval-augmented language model serving
  with speculation.
\newblock In {\em Forty-first International Conference on Machine Learning}.

\bibitem{zhao-etal-2024-optimizing}
Yiyun Zhao, Prateek Singh, Hanoz Bhathena, Bernardo Ramos, Aviral Joshi,
  Swaroop Gadiyaram, and Saket Sharma.
\newblock Optimizing {LLM} based retrieval augmented generation pipelines in
  the financial domain.
\newblock In Yi~Yang, Aida Davani, Avi Sil, and Anoop Kumar, editors, {\em
  Proceedings of the 2024 Conference of the North American Chapter of the
  Association for Computational Linguistics: Human Language Technologies
  (Volume 6: Industry Track)}, pages 279--294, Mexico City, Mexico, June 2024.
  Association for Computational Linguistics.

\bibitem{zheng2024sglang}
Lianmin Zheng, Liangsheng Yin, Zhiqiang Xie, Chuyue Sun, Jeff Huang, Cody~Hao
  Yu, Shiyi Cao, Christos Kozyrakis, Ion Stoica, Joseph~E. Gonzalez, Clark
  Barrett, and Ying Sheng.
\newblock {SGL}ang: Efficient execution of structured language model programs.
\newblock In {\em The Thirty-eighth Annual Conference on Neural Information
  Processing Systems}, 2024.

\bibitem{qmsum}
Ming Zhong, Da~Yin, Tao Yu, Ahmad Zaidi, Mutethia Mutuma, Rahul Jha, Ahmed
  Hassan~Awadallah, Asli Celikyilmaz, Yang Liu, Xipeng Qiu, and Dragomir Radev.
\newblock {QMS}um: {A} {N}ew {B}enchmark for {Q}uery-based {M}ulti-domain
  {M}eeting {S}ummarization.
\newblock In {\em North American Association for Computational Linguistics
  (NAACL)}, 2021.

\bibitem{distserve}
Yinmin Zhong, Shengyu Liu, Junda Chen, Jianbo Hu, Yibo Zhu, Xuanzhe Liu, Xin
  Jin, and Hao Zhang.
\newblock {DistServe}: Disaggregating prefill and decoding for
  goodput-optimized large language model serving.
\newblock In {\em 18th USENIX Symposium on Operating Systems Design and
  Implementation (OSDI 24)}, pages 193--210, Santa Clara, CA, July 2024. USENIX
  Association.

\bibitem{zhou2025indepthanalysisgraphbasedrag}
Yingli Zhou, Yaodong Su, Youran Sun, Shu Wang, Taotao Wang, Runyuan He, Yongwei
  Zhang, Sicong Liang, Xilin Liu, Yuchi Ma, and Yixiang Fang.
\newblock In-depth analysis of graph-based rag in a unified framework, 2025.

\bibitem{nanoflow}
Kan Zhu, Yilong Zhao, Liangyu Zhao, Gefei Zuo, Yile Gu, Dedong Xie, Yufei Gao,
  Qinyu Xu, Tian Tang, Zihao Ye, et~al.
\newblock Nanoflow: Towards optimal large language model serving throughput.
\newblock {\em arXiv preprint arXiv:2408.12757}, 2024.

\bibitem{zhu2024ragevalscenariospecificrag}
Kunlun Zhu, Yifan Luo, Dingling Xu, Ruobing Wang, Shi Yu, Shuo Wang, Yukun Yan,
  Zhenghao Liu, Xu~Han, Zhiyuan Liu, and Maosong Sun.
\newblock Rageval: Scenario specific rag evaluation dataset generation
  framework, 2024.

\end{thebibliography}

\appendix
\section{Appendix}
\label{sec:appendix}

The appendix has not been not peer-reviewed.

\lstset{
    basicstyle=\ttfamily\small, 
    keywordstyle=\color{blue}, 
    commentstyle=\color{gray}, 
    stringstyle=\color{red},  
    numbers=left,            
    numberstyle=\tiny\color{gray}, 
    frame=single,            
    breaklines=true          
}

\subsection{Prompt and input syntax for \name' LLM profiler} 
We use a simple prompt to provide the metadata to \name' LLM profiler. We don't perform any prompt tuning or optimizations for this work as the goal of the prompt was only to get binary decisions from the natural language properties of the query.

\begin{lstlisting}[language=Python]

f"""
For the given query = {get.query()}: Analyse the language and internal structure of the query and provide the following information :

    1. Does it needs joint reasoning across multiple documents or not.
    2. Provide a complexity profile for the query: 
        Complexity: High/Low \n \
        Joint Reasoning needed: Yes/No \n "
    3. Does this query need input chunks to be summarized and if yes, provide a range in words for the summarized chunks.
    4. How many pieces of information is needed to answer the query?

    database_metadata = {get.metadata()}
    chunk_size = {get.chunk_size()}
        
Estimate the query profile along with the database_metadata and chunk_size to provide the output.

"""
\end{lstlisting}

The metadata is a single line summary of the content of the database. For example, for KG RAG FinSec , the metadata is derived from the dataset definition.

\begin{figure}[t]
\begin{lstlisting}[language=Python]

def get_metadata():

    metadata = "The dataset consists of multiple chunks of information from Fortune 500 companies on financial reports from every quarter of 2023. The chunk size is 1024 tokens."
    
    return metadata

\end{lstlisting}

\end{figure}

The \chunksize is chosen based on guidelines RAG literature for different types of RAG tasks~\cite{ragblog1,ragblog2}. We don't tune this knob as it is fixed when the database is created. Finally in this work, we don't tune the metadata for the dataset, we use the existing summaries. 

Today, RAG QA datasets already have summaries present along with the queries and contexts. In future work , it will be interesting to study how to effectively construct such a metadata for newer datasets. One possible solution could be an LLM summarizer on a set of values from the dataset which opens up further avenues to perform joint scheduling and configuration tuning.

\subsection{Changing the embedding algorithm for the vector database in \name }

\name picks a state-of-art retrieval algorithm \textit{Cohere-embed-v3.0}~\cite{cohere_embed}. Using two other popular retrieval algorithms \textit{All-mpnet-base-v2}~\cite{reimers2019sentencebertsentenceembeddingsusing} and \textit{text-embedding-3-large-256}~\cite{neelakantan2022textcodeembeddingscontrastive}, the F1-score change remained within 1\%. The delay has no measurable difference as the retrieval is $> 100\times$ faster than LLM synthesis~\cite{acl_rag_tutorial}.

\end{document}